\documentclass{article}%
\usepackage{amsmath}
\usepackage{amsfonts}
\usepackage{amssymb}
\usepackage{graphicx}
\usepackage{url,hyperref}
\usepackage{algorithmic}
\usepackage{algorithm}%
\providecommand{\U}[1]{\protect \rule{.1in}{.1in}}

\begin{document}

\title{Survival Analysis as Imprecise Classification with Trainable Kernels}
\author{Andrei V. Konstantinov, Vlada A. Efremenko and Lev V. Utkin\\Higher School of Artificial Intelligence Technologies\\Peter the Great St.Petersburg Polytechnic University\\St.Petersburg, Russia\\e-mail: konstantinov\_av@spbstu.ru, efremenko\_va@spbstu.ru, \\utkin\_lv@spbstu.ru}
\date{}
\maketitle

\begin{abstract}
Survival analysis is a fundamental tool for modeling time-to-event data in
healthcare, engineering, and finance, where censored observations pose
significant challenges. While traditional methods like the Beran estimator
offer nonparametric solutions, they often struggle with the complex data
structures and heavy censoring. This paper introduces three novel survival
models, iSurvM (the \textbf{i}mprecise \textbf{Surv}ival model based on
\textbf{M}ean likelihood functions), iSurvQ (the \textbf{i}mprecise
\textbf{Surv}ival model based on the \textbf{Q}uantiles of likelihood
functions), and iSurvJ (the \textbf{i}mprecise \textbf{Surv}ival model based
on the \textbf{J}oint learning), that combine imprecise probability theory
with attention mechanisms to handle censored data without parametric
assumptions. The first idea behind the models is to represent censored
observations by interval-valued probability distributions for each instance
over time intervals between events moments. The second idea is to employ the
kernel-based Nadaraya-Watson regression with trainable attention weights for
computing the imprecise probability distribution over time intervals for the
entire dataset. The third idea is to consider three decision strategies for
training, which correspond to the proposed three models. Experiments on
synthetic and real datasets demonstrate that the proposed models, especially
iSurvJ, consistently outperform the Beran estimator from the accuracy and
computational complexity points of view. Codes implementing the proposed
models are publicly available.

\textit{Keywords}: survival analysis, censored observations, classification,
attention mechanism, imprecise probabilities

\end{abstract}

\section{Introduction}

Survival analysis \cite{Hosmer-Lemeshow-May-2008}, or time-to-event analysis,
provides a robust framework for modeling time-to-event data across diverse
domains. In healthcare, it predicts critical outcomes such as mortality,
disease recurrence, and recovery timelines while quantifying treatment
effectiveness and disease progression. In engineering, the method assesses
reliability, predicting system failures, optimizing maintenance schedules, and
evaluating product durability. It also proves valuable in financial risk
modeling (e.g., credit defaults and customer churn) and industrial contexts
(e.g., equipment degradation and failure prediction). A key strength of
survival analysis is its ability to handle censored data, making it
indispensable for scenarios with incomplete event observations, from clinical
trials to mechanical system monitoring. This approach delivers quantifiable
insights into event timing uncertainty across these fields.

Due to the importance of tasks addressed by survival analysis, numerous
machine learning models have been developed to handle time-to-event data and
solve related problems within this framework
\cite{jing2019deep,hao2021deep,Lee-Wang-2003,Hothorn-etal-2006,Wrobel-etal-2017,Zhao-Feng-2019,zhao2024tutorial}%
. Comprehensive reviews of these models are available in
\cite{marinos2021survey,Wang-Li-Reddy-2019,Wiegrebe:2024aa}.

Many of these models adapt and extend traditional classification and
regression techniques to handle the unique challenges of survival analysis,
for example, random survival forests \cite{Ishwaran-Kogalur-2007}, survival
support vector machines \cite{Belle-etal-2008}, neural networks
\cite{chen2024introduction}, survival-oriented transformers
\cite{arroyo2024deep,hu2021transformer,Li-Zhu-Yao-Huang-22,zhang2025adaptive}.
These methods demonstrate the growing intersection of classical survival
analysis with modern machine learning paradigms.

Survival analysis often involves modeling time-to-event data in the presence
of covariates or feature vectors. While traditional nonparametric methods like
the Kaplan-Meier estimator \cite{Kaplan-Meier-58} assume independence between
survival times and covariates, conditional survival models incorporate
covariate information to provide more personalized estimates. A prominent
example is the Beran estimator \cite{Beran-81}, a nonparametric conditional
survival function estimator that extends the Kaplan-Meier approach by
incorporating kernel weighting based on covariates. Kernel-based methods, like
the Beran estimator, provide flexible, nonparametric approaches to survival
analysis by leveraging smoothing techniques to model time-to-event data with
covariates. They offer a powerful toolkit for survival analysis, especially
when parametric assumptions fail. Chen \cite{chen2024survival} presented a
neural network framework based on kernels. The deep kernel conditional
Kaplan-Meier estimator was introduced in \cite{chen2024survival}. Sparse
kernel methods based on applying SVM were considered in \cite{evers2008sparse}%
. Instead of estimating the survival function directly, some approaches smooth
the cumulative hazard function \cite{gefeller1992review}. A series of survival
models based on applying Bayesian kernel methods which combine the flexibility
of kernel-based nonparametric modeling with the probabilistic rigor of
Bayesian inference \cite{cawley2004bayesian}. Another kernel-based method is
the kernel Cox regression \cite{li2002kernel,rong2024kernel,yang2021weighted}
which relaxes the assumption of the linear covariate effects accepted in the
original Cox model \cite{Cox-1972} by allowing nonlinear effects via kernel functions.

Survival analysis is traditionally framed as a time-to-event prediction
problem, but it can also be approached as a classification task by
discretizing time and predicting event probabilities over intervals in the
framework of discrete-time survival analysis \cite{tutz2016modeling}, where
continuous survival data can be converted into a person-period format
\cite{suresh2022survival}. The loss functions in this case are based on
applying the likelihood function \cite{kvamme2021continuous}. A method for
treating survival analysis as a classification problem was proposed in
\cite{zhong1909survival} where a \textquotedblleft stacking\textquotedblright%
\ idea is used to collect features and outcomes of the survival data in a
large data frame, and then to solve the classification problem.

We propose a novel approach within the framework of discrete-time survival
analysis, which can be characterized as imprecise survival modeling. The core
idea is to model non-parametric probability distributions over time intervals
for each instance in the training set. For uncensored observations, the
distribution is degenerate, where the event's true time interval has
probability one while all others have probability zero. For censored
observations, the probability of the event occurring in any interval after the
censored time can range from zero to one, reflecting prior ignorance. This
introduces imprecision, meaning each instance induces a set of possible
probability distributions. To aggregate the probability distributions
generated for each instance from the training set, we propose to employ kernel
Nadaraya-Watson regression \cite{Nadaraya-1964,Watson-1964} with trainable
dot-product attention weights \cite{Luong-etal-2015,Vaswani-etal-17}. We
propose three training strategies, each leading to a distinct model. The first
two strategies are based on generating random probability distributions from
the produced sets of distributions. The first strategy averages all loss
functions in the form of the likelihood function in accordance with the
aggregated probability distributions. The corresponding model is called
\emph{iSurvM} (the \textbf{i}mprecise \textbf{Surv}ival model based on
\textbf{M}ean likelihood functions). The second strategy averages only a
portion of the largest values of the loss functions over the generated
probability distributions. The corresponding model is called \emph{iSurvQ}
(the \textbf{i}mprecise \textbf{Surv}ival model based on the \textbf{Q}%
uantiles of likelihood functions). According to the third strategy, the
probability distributions and the attention weights are jointly learned in a
special way without using the generation procedure. The corresponding model is
called \emph{iSurvJ} (the \textbf{i}mprecise \textbf{Surv}ival model based on
the \textbf{J}oint learning). We avoid any assumptions concerning with a
survival model, for example, with the Cox model, which define the loss
functions for training the models (see, for example, \cite{Kvamme-etal-19}).

In contrast to many imprecise models for survival analysis
\cite{Coolen-97,Coolen-Yan-2004,Mangili-etal-2015}, the proposed models do not
consider the incomplete or partial knowledge in event time data. They
represent censored observations as interval-valued or imprecise data.

Our contributions can be summarized as follows:

\begin{enumerate}
\item We propose three training strategies and the corresponding survival
models, iSurvM, iSurvQ, iSurvJ, as an alternative to kernel-based models such
as the Beran estimator.

\item The models can be trained using different attention mechanisms
implemented by means of neural networks as well as simple Gaussian kernels (an
additional model \emph{iSurvJ(G)} which uses the Gaussian kernel in the
attention weights). The number of trainable parameters depends on the key and
query attention matrices and can be arbitrarily chosen.

\item Additionally, unlike Beran?s estimators, the proposed models impose no restrictions on the number of concurrent event times, making them more versatile. Moreover, a high proportion of censored data does not lead to significant accuracy degradation in the proposed models, unlike in Beran?s estimators.

\item No parametric assumptions are made in the proposed models.

\item Various numerical experiments with real and synthetic datasets are
conducted to compare the proposed survival models each other under different
conditions and to compare them with the Beran estimator using the concordance
index and the Brier score. We compare the proposed models with the Beran
estimator, but not with available transformer-based models
\cite{hu2021transformer,tang2023explainable,wang2024resdeepsurv}, because the
introduced models are regarded as alternatives to the kernel-based Beran
estimator and can be incorporated into a more complex model as an component.
The corresponding codes implementing the proposed models are publicly
available at: \url{https://github.com/NTAILab/iSurvMQJ}.
\end{enumerate}

The paper is organized as follows. Related work devoted to machine learning
models in survival analysis, to kernel-based methods and models in survival
analysis, to imprecise models in survival analysis can be found in Section 2.
Section 3 provides basic definitions of survival analysis, the attention
mechanism and the Nadaraya-Watson regression. Survival analysis as an
imprecise multi-label classification problem is considered in Section 4.
Training strategies and survival models iSurvM, iSurvM, iSurvM are studied in
Section 5. Numerical experiments with well-known public real and synthetic
data illustrating the proposed models are given in Section 6 and in the
Appendix section. Concluding remarks are provided in Section 7.

\section{Related work}

\textbf{Machine learning models in survival analysis}. Survival analysis,
which deals with time-to-event data, has seen significant advancements through
the integration of machine learning techniques. Numerous models have been
developed to predict survival times, hazard functions, and other key measures,
leveraging both traditional statistical methods and modern computational
approaches. Several comprehensive reviews and surveys provide detailed
insights into these developments. For instance, \cite{Wang-Li-Reddy-2019}
offer a broad examination of machine learning methods in survival analysis,
while \cite{marinos2021survey} and \cite{salerno2023high} discuss recent
trends, including deep learning and high-dimensional data applications.
Additionally, \cite{bender2020general} provide a methodological perspective on
generalizing survival models, and \cite{EmmertStreib-Dehmer-19} explore
interpretability and benchmarking in survival machine learning. A more
contemporary discussion on emerging techniques and challenges in the field can
be found in \cite{Wiegrebe:2024aa}.

\textbf{Kernel-based methods and models in survival analysis.} An important
kernel-based approach to survival analysis is the Beran estimator
\cite{Beran-81}, which extends the Kaplan-Meier estimator by incorporating
covariate information through kernel smoothing. The Beran estimator or the
conditional Kaplan-Meier estimator \cite{chen2020deep} enables robust
estimation of survival functions in complex, high-dimensional settings. Recent
advances have further enhanced these techniques through integration with deep
learning. For instance, \cite{chen2024survival} proposed a neural network
framework that automatically learns optimal kernel functions for survival
analysis, leading to the development of the deep kernel conditional
Kaplan-Meier estimator, which improves adaptability to heterogeneous data structures.

Alternative kernel-based strategies include sparse kernel methods via support
vector machines (SVMs) \cite{evers2008sparse}, which enhance computational
efficiency, and approaches that focus on smoothing the cumulative hazard
function rather than the survival function directly \cite{gefeller1992review}.
Bayesian formulations of kernel methods, such as those in
\cite{cawley2004bayesian}, combine the flexibility of nonparametric modeling
with the probabilistic rigor of Bayesian inference, offering uncertainty
quantification alongside predictive accuracy.

Another significant direction is the kernel Cox regression \cite{li2002kernel,
rong2024kernel, yang2021weighted}, which generalizes the classical Cox model
\cite{Cox-1972} by replacing linear covariate effects with nonlinear
kernel-based relationships. This relaxation captures intricate dependencies in
the data, addressing a key limitation of the original proportional hazards
framework. Collectively, these methods demonstrate the versatility of
kernel-based techniques in survival analysis, balancing interpretability,
computational tractability, and adaptability to diverse data regimes.

\textbf{Attention mechanism}. Recent advances in survival analysis have
leveraged attention mechanisms \cite{Vaswani-etal-17} to improve
interpretability and predictive accuracy. Deep learning approaches use
attention to model complex, non-linear relationships. The corresponding
methods address key challenges like censoring, competing risks, and
non-proportional hazards while providing interpretable feature importance. The
attention mechanism was applied to different deep learning survival models. In
particular, a self-attention mechanism was used to model local and global
context simultaneously for survival analysis in \cite{yang2024deep}. The
attention mechanism can be a part of transformers which capture long-range
dependencies in survival data, offering improved performance in
high-dimensional settings. Transformers for survival analysis with competing
events were proposed in \cite{Wang-Sun-22}. Transformers solving survival
machine learning problems were also proposed in
\cite{hu2021transformer,tang2023explainable,wang2024resdeepsurv}. Different
applications of transformers to medical problems in the framework of survival
analysis can be found in
\cite{jiang2023mhattnsurv,teng2025semi,wang2023surformer,yao2024multi}.

\textbf{Imprecise models in survival analysis}. It should be noted that the
imprecise survival analysis extends classical survival models to account for
partial or incomplete knowledge in time-to-event data. Unlike traditional
methods that assume precise probabilities, imprecise models incorporate
set-valued or interval-based estimates to reflect epistemic uncertainty,
ambiguity in censoring mechanisms, or conflicting expert opinions. A method of
statistical inference based on the Hill model \cite{hill1988finetti} for data
that include right-censored observations was proposed in
\cite{Coolen-Yan-2004} where the authors derive bounds for the predictive
survival function. Another method of statistical inference based on the
imprecise Dirichlet model \cite{Walley96a} was proposed in \cite{Coolen-97}. A
method applying a robust Dirichlet process for estimating survival functions
from samples with right-censored data was introduced in
\cite{Mangili-etal-2015}. It should be note that the proposed imprecise models
cope with cases when exact event times are unknown or are interval-valued and
strong distributional assumptions are violated.

\section{Preliminaries}

\subsection{Survival analysis}

Datasets in survival analysis are represented by a set of triplets
$\mathcal{A}=\{(\mathbf{x}_{1},\delta_{1},T_{1}),...,(\mathbf{x}_{N}%
,\delta_{N},T_{N})\}$. Here the vector $\mathbf{x}_{i}\in \mathbb{R}^{d}$
consisting of $d$ features characterizes the $i$-th object. The time $T_{i}$
corresponds to one of two types of the event observations. The first type is
when the event is observed. In this case, the observation is called uncensored
and the censoring indicator $\delta_{i}$ is equal to $1$. The second type is
when the event is not observed, it is greater than $T_{i}$, but its `true'
value is unknown. In this case, the observation is called censored and the
censoring indicator $\delta_{i}$ is equal to $0$. A survival model is trained
on the set $\mathcal{A}$ to predict probabilistic measures of an event time
$T$ for a new object representing by the vector $\mathbf{x}$.

Important concepts in survival analysis are the survival function (SF) and the
cumulative hazard function (CHF). The SF $S(t\mid \mathbf{x})$ is a function of
time $t$ defined as the probability of surviving up to time $t$, i.e.,
$S(t\mid \mathbf{x})=\Pr \{T>t\mid \mathbf{x}\}$. The CHF $H(t\mid \mathbf{x})$ is
also a function of time defined through the SF as follows:
\begin{equation}
H(t\mid \mathbf{x})=-\ln S(t\mid \mathbf{x}). \label{RSF_Arr_10}%
\end{equation}

Many survival machine learning models have been developed in the last decades.
In order to compare the models, special measures are used different from the
standard accuracy measures accepted in machine learning classification and
regression models. The most popular measure in survival analysis is Harrell's
C-index (concordance index) \cite{Harrell-etal-1982}. It estimates the
probability that, in a randomly selected pair of objects, the object that
fails first had a worst predicted outcome. In fact, this is the probability
that the event times of a pair of objects are correctly ranking. C-index does
not depend on choosing a fixed time for evaluation of the model and takes into
account censoring of patients \cite{May-etal-2004}.

Let a training set $\mathcal{A}$ consist of $n$ triplets $(\mathbf{x}%
_{i},\delta_{i},T_{i})$, $i=1,...,N$. Survival analysis aims to estimate the
event time $T$ for a new instance $\mathbf{x}$ on the basis of the dataset
$\mathcal{A}$. Important concepts in survival analysis are survival functions
(SFs) and cumulative hazard functions (CHFs). The conditional SF denoted as
$S(t\mid \mathbf{x})$ is the probability of surviving up to time $t$, that is
$S(t\mid \mathbf{x})=\Pr \{T>t\mid \mathbf{x}\}$.

Comparison of survival models is often carried out by means of the C-index
\cite{Harrell-etal-1982}. It estimates the probability that the event times of
a pair of instances are correctly ranking. Let $\mathcal{J}$ is a set of all
pairs $(i,j)$ satisfying conditions $\delta_{i}=1$ and $T_{i}<T_{j}$. The
C-index is formally computed as \cite{Uno-etal-11,Wang-Li-Reddy-2019}:%
\begin{equation}
C=\frac{\sum_{(i,j)\in \mathcal{J}}\mathbf{1}[\widehat{T}_{i}<\widehat{T}_{j}%
]}{\sum_{(i,j)\in \mathcal{J}}1}, \label{Survival_DF_24}%
\end{equation}
where $\widehat{T}_{i}$ and $\widehat{T}_{j}$ are predicted expected event times.

If the C-index is equal to 1, then the corresponding survival model is
supposed to be perfect. If the C-index is 0.5, then the model is not better
than random guessing.

Another index used to compare survival models us the Brier Score (BS)
\cite{Brier-1950,Graf-etal-99} which is defined as
\begin{equation}
BS(t,\hat{S})=\mathbb{E}\left[  \left(  \Delta_{new}(t)-\hat{S}(t\mid
\mathbf{x}_{new})\right)  ^{2}\right]  ,
\end{equation}
where $\Delta_{new}(t)=\mathbf{1}\left(  T_{new}>t\right)  $ where is the true
status of a new test subject and $\hat{S}(t\mid \mathbf{x}_{new})$ is the
predicted survival probability.

The Integrated Brier Score (IBS) extends the BS by averaging it over a range
of time points, typically from $t=0$ to a maximum time $t_{\max}$. It is
defined as:
\begin{equation}
IBS=\frac{1}{t_{\max}}\int_{0}^{t_{\max}}BS(t,\hat{S})\,dt.
\end{equation}

\subsection{Attention mechanism and the Nadaraya-Watson regression}

The idea of the attention mechanism can be clearly illustrated using the
Nadaraya-Watson kernel regression model \cite{Nadaraya-1964,Watson-1964}.
Consider a training set $\{(\mathbf{x}_{1},y_{1}),...,(\mathbf{x}_{N}%
,y_{N})\}$ with $N$ instances, where each $\mathbf{x}_{i}\in \mathbb{R}^{d}$ is
a feature vector and $y_{i}\in \mathbb{R}$ is its corresponding label. For a
new input feature vector $\mathbf{x}_{0}$, the regression output prediction
$\widetilde{y}_{0}$ can be estimated as a weighted average using the
Nadaraya-Watson kernel regression model \cite{Nadaraya-1964,Watson-1964}:%
\begin{equation}
\widetilde{y}_{0}=\sum_{i=1}^{N}a_{0,i}(\mathbf{w})y_{i}.
\end{equation}

Here $a_{0,i}(\mathbf{w})$ represents the attention weight with trainable
parameters $\mathbf{w}$, which quantifies the similarity (or distance) between
the input feature vector $\mathbf{x}_{0}$ and the training feature vector
$\mathbf{x}_{i}$. The closer $\mathbf{x}_{0}$ to $\mathbf{x}_{i}$, the larger
the corresponding weight $a_{0,i}(\mathbf{w})$ becomes. In general, any
distance or similarity function satisfying this monotonicity condition can
serve as attention weights. A natural choice is the family of kernel
functions, since a kernel $K$ inherently acts as a similarity measure between
vectors $\mathbf{x}_{i}$ and $\mathbf{x}_{0}$. Therefore, the attention
weights can be expressed as:
\begin{equation}
a_{0,i}(\mathbf{w})=\frac{K(\mathbf{x}_{0},\mathbf{x}_{i},\mathbf{w})}%
{\sum_{j=1}^{N}K(\mathbf{x}_{0},\mathbf{x}_{j},\mathbf{w})}.
\label{attent_weights}%
\end{equation}

In the context of the attention mechanism \cite{Bahdanau-etal-14}, the vector
$\mathbf{x}_{0}$ is referred to as the \textit{query}, while vectors
$\mathbf{x}_{i}$ and labels $y_{i}$ are called the \textit{keys} and
\textit{values}, respectively. The attention weights $a_{0,i}(\mathbf{w})$ can
be generalized by introducing trainable parameters. For example, using a
Gaussian kernel with trainable parameter vector $\mathbf{w}=(w_{1},\dots
,w_{n})$, the attention weight can be expressed as:
\begin{align}
a_{0,i}(\mathbf{w})  &  =\sigma \left(  -\left \Vert \mathbf{x}_{0}%
-\mathbf{x}_{i}\right \Vert ^{2}|\mathbf{w}\right) \nonumber \\
&  =\frac{\exp \left(  -w_{i}\left \Vert \mathbf{x}_{0}-\mathbf{x}%
_{i}\right \Vert ^{2}\right)  }{\sum_{j=1}^{N}\exp \left(  -w_{i}\left \Vert
\mathbf{x}_{0}-\mathbf{x}_{i}\right \Vert ^{2}\right)  },
\end{align}
where $\sigma(\cdot)$ is the softmax function.

Several definitions of attention weights and corresponding attention
mechanisms exist in the literature. Notable examples include the additive
attention \cite{Bahdanau-etal-14} and multiplicative or dot-product attention
\cite{Luong-etal-2015,Vaswani-etal-17}.

\section{Survival analysis as an imprecise multi-label classification problem}

This section demonstrates how survival models can be formulated using the
Nadaraya-Watson kernel regression.

Consider a partition of the time axis into discrete intervals:
\begin{equation}
\lbrack0,\infty)=[0,t_{1}]\cup \lbrack t_{1},t_{2}]\cup...\cup \lbrack
t_{T-2},t_{T-1}]\cup \lbrack t_{T-1},\infty),
\end{equation}
such that
\begin{equation}
0=t_{0}<t_{1}<t_{2}<...<t_{T-1}<\infty.
\end{equation}

The partition divides the time axis into $T$ intervals denoted as $\tau
_{i}=[t_{i-1},t_{i}]$ for $i=1,\dots,T$. We assume that at each time point
$t_{i}$ (or within interval $\tau_{i}$), there are $u_{i}$ uncensored events
and $c_{i}$ right-censored events such that the total number of events is
$N=\sum_{i=1}^{T}(u_{i}+c_{i})$.

Let us define the probability $\pi_{k}^{(i)}$ that an event corresponding to
the $i$-th object is observed in the interval $\tau_{k}$. This generates a
probability distribution for each subject $\mathbf{\pi}^{(i)}=(\pi_{1}%
^{(i)},...,\pi_{T}^{(i)})$, $i=1,...,N$. If the corresponding observation is
uncensored, then there holds%
\begin{equation}
\pi_{j}^{(i)}=\left \{
\begin{array}
[c]{cc}%
0, & j\neq k,\\
1, & j=k.
\end{array}
\right.  \label{uncensor_distr}%
\end{equation}

For censored observations, the true event time must occur in some interval
after $t_{k-1}$, though the specific interval remains unknown. Consequently,
the probability that the event is at each interval after the time $t_{k-1}$
can be from $0$ to $1$, i.e., there holds%
\begin{equation}
\pi_{j}^{(i)}=\left \{
\begin{array}
[c]{cc}%
0, & j<k,\\
\lbrack0,1], & j\geq k.
\end{array}
\right.  \label{censor_distr}%
\end{equation}

The interval $[0,1]$ is a form of prior ignorance. In all cases, we have the
constraint $\sum_{j=1}^{T}\pi_{j}^{(i)}=1$ for precise probabilities. However,
this condition does not fulfil when the upper or lower probabilities are
summed. Due to imprecision of probabilities $\pi_{j}^{(i)}$, we suppose that
the $i$-th instance produces a set of probability distributions $\mathbf{\pi
}^{(i)}$ denoted as $\mathcal{R}^{(i)}$.

Let us consider the following \emph{classification problem}. There is a set of
feature vectors $\mathbf{x}_{1},...,\mathbf{x}_{N}$. The $k$-th interval
$[t_{k-1},t_{k}]$ can be viewed as the $k$-th class represented by the vector
$\mathbf{\pi}_{k}=(\pi_{k}^{(1)},...,\pi_{k}^{(N)})^{\mathrm{T}}$. Then the
Nadaraya-Watson kernel regression can be applied to find the class probability
distribution $\mathbf{p}(\mathbf{x}_{0})=(p_{1}(\mathbf{x}_{0}),...,p_{T}%
(\mathbf{x}_{0}))$ of a new instance $\mathbf{x}_{0}$. Since a part of
probabilities $\pi_{k}^{(i)}$ are interval values, then it is obvious that the
predicted class probabilities $p_{1},...,p_{T}$ are also interval-valued. They
are determined as follows:%
\begin{equation}
p_{k}(\mathbf{x}_{0})=\sum_{i=1}^{N}a_{0,i}(\mathbf{w})\pi_{k}^{(i)}%
,\ k=1,...,T, \label{N-W-2}%
\end{equation}
where $a_{0,i}(\mathbf{w})$ is an attention weight expressed through kernels
$K(\mathbf{x}_{0},\mathbf{x}_{i},\mathbf{w})$ (see (\ref{attent_weights})),
$i=1,...,N$; $\mathbf{w}=(w_{1},...,w_{T})$ is the trainable vector of parameters.

The attention weights satisfy the condition%
\begin{equation}
\sum_{i=1}^{N}a_{0,i}(\mathbf{w})=1.
\end{equation}

\section{Training strategies and three survival models}

\subsection{A general form of attention weights}

Before presenting the proposed survival models, their common elements
concerning the attention weights and regularization will be described. We
consider the attention weights $a_{0,i}(\mathbf{w})$ in a general form. Let
$\mathbf{K}\in \mathbb{R}^{N\times d_{0}}$ represent the matrix of the input
keys (the matrix of $N$ feature vectors $\mathbf{x}_{i}$, $i=1,...,N$, from
the training set), and $\mathbf{Q}\in \mathbb{R}^{N\times d_{0}}$ represent the
matrix of the input queries (the matrix of the same feature vectors from the
training set). Here $d_{0}$ is the dimension of the initial feature vectors
$\mathbf{x}_{i}$, $i=1,...,N$. The matrix $\mathbf{Q}$ is used in the training
phase. In many tasks, the sparse representation of feature vectors might be
useful. In order to implement it, matrices $\mathbf{K}$ and $\mathbf{Q}$ can
be transformed to sets of vectors of a new dimension $d\geq d_{0}$ by using a
neural network with parameters $\mathbf{\theta}$ implementing functions
$f_{\mathbf{\theta}}(\mathbf{K})$ and $f_{\mathbf{\theta}}(\mathbf{Q})$ as
\begin{equation}
\mathbf{K}=f_{\mathbf{\theta}}(\mathbf{K}_{0})\in \mathbb{R}^{N\times d}%
,\quad \mathbf{Q}=f_{\mathbf{\theta}}(\mathbf{Q}_{0})\in \mathbb{R}^{N\times d},
\label{Dimension1}%
\end{equation}
where $f_{\mathbf{\theta}}$ is a parameterized neural network trained with
weights $\mathbf{\theta}$.

Let $\mathbf{W}_{K}\in \mathbb{R}^{d\times d}$ and $\mathbf{W}_{Q}\in
\mathbb{R}^{d\times d}$ be the trainable matrices. Then the matrix
$\mathbf{A}\in \mathbb{R}^{N\times N}$ of attention weights $a_{ij}$,
$i=1,...,N$, $j=1,...,N$, can be computed as
\cite{Luong-etal-2015,Vaswani-etal-17}:
\begin{equation}
\mathbf{A=}\frac{\left(  \mathbf{QW}_{Q}\right)  \left(  \mathbf{KW}%
_{K}\right)  }{\sqrt{d}}. \label{Attention_1}%
\end{equation}

Each element $a_{ij}$ of $\mathbf{A}$ represents the similarity between the
$i$-th query and $j$-th key. Matrices $\mathbf{W}_{Q}$ and $\mathbf{W}_{K}$
are randomly initialized in the training phase. The attention weights are also
represented as a neural network with parameters $\mathbf{W}_{Q}$ and
$\mathbf{W}_{K}$.

To restrict certain interactions between elements of matrix $\mathbf{A}$, we
apply an attention mask $\mathbf{M}\in \{-\infty,1\}^{N\times N}$ which is
initialized with uniform random values $m_{ij}\sim U(0,1)$. Then it is
binarized using a threshold $p_{mask}$ (a hyperparameter) as follows:
\begin{equation}
m_{ij}=%
\begin{cases}
1, & m_{ij}>p_{mask},\\
-\infty, & \text{otherwise}.
\end{cases}
,\ m_{ii}=-\infty,\  \forall i=1,...,N \label{Attention_mask}%
\end{equation}

The mask is applied via element-wise multiplication: $\mathbf{A}%
=\mathbf{A}\odot \mathbf{M}$. The row-wise softmax normalization yields the
final attention weights:%

\begin{equation}
w_{i,j}=\frac{\exp(a_{i,j})}{\sum_{k}\exp(a_{i,k})}, \label{Weight_matrix_W}%
\end{equation}
which comprise the final attention weight matrix $\mathbf{W}\in \mathbb{R}%
^{N\times N}$.

Hence, (\ref{N-W-2}) can be written in a more general form:
\begin{equation}
p_{k}(\mathbf{x}_{0})=\mathbf{W\pi}_{k},\ k=1,...,T, \label{class_prob_distr}%
\end{equation}

In order to solve the above classification problem under condition of the
interval-valued class probabilities, we have to define how to train the
classifier when the class probabilities are interval-valued.

Let $\Pi=(\mathbf{\pi}^{(1)},...,\mathbf{\pi}^{(N)})$ and $\mathcal{R}%
=\mathcal{R}^{(1)}\cup...\cup \mathcal{R}^{(N)}$. Introduce the loss function
for solving the classification problem $\mathcal{L}(\Theta,\mathbf{\Pi})$,
where $\Theta$ is a set of training parameters, which includes parameters
$\mathbf{W}_{K}$ and $\mathbf{W}_{Q}$ from (\ref{Attention_1}),
$\mathbf{\theta}$ from (\ref{Dimension1}). One of the ways for dealing with
the loss function under condition of the set of probabilities is to consider
the following optimization problem:
\begin{equation}
\min_{\Theta}\mathbb{E}_{\Pi \in \mathcal{R}}\mathcal{L}(\Theta,\mathbf{\Pi}).
\end{equation}

To handle the above optimization problem, we apply the Monte Carlo sampling
scheme. The expectation is replaced with the sum of objective functions over a
set of different probability distributions $\mathbf{\pi}^{(1)},...,\mathbf{\pi
}^{(N)}$ taken from $\mathcal{R}^{(1)},...,\mathcal{R}^{(N)}$, respectively.
It can be done by generating $M$ discrete probability distributions
$\mathbf{q}_{1}^{(i)},...,\mathbf{q}_{M}^{(i)}$ for the $i$-th instance from
$\mathcal{R}^{(i)}$. Here $M$ is a hyperparameter. If the $i$-th observation
is uncensored ($\delta=1$), then the probability distribution $\mathbf{\pi
}^{(i)}$ is degenerate having the unit value at the interval of time where the
corresponding event occurred (see (\ref{uncensor_distr})). If the $i$-th
observation is censored ($\delta=0$), then probabilities are zero before the
censoring interval. Probabilities of the subsequent intervals of time are
interval-valued (see (\ref{censor_distr})). The corresponding distributions
are randomly generated via the Dirichlet distribution
\cite{Rubinstein-Kroese2008}. There are also different approaches to generate
points from the unit simplex \cite{Smith-Tromble-04}. As a result, we obtain a
set of $N$ matrices $\mathbf{S}^{(i)}\in \mathbb{R}^{M\times T}$, $i=1,...,N$,
such that each matrix contains $M$ probability distributions $\mathbf{q}%
_{1}^{(i)},...,\mathbf{q}_{M}^{(i)}$ from $\mathcal{R}^{(i)}$. It should be
noted that we generate only distributions corresponding to interval-valued
probabilities $\pi_{j}^{(i)}$. This implies that the distributions are
generated only for censored observations, the unit simplex dimension depends
on the time interval, where the event occurs.

Let $c(i)$ be the index of the time interval in which the event, corresponding
to the $i$-th instance, occurs. Then the unit simplex dimension for generating
a distribution is equal to $T-c(i)$.

Hence, the attention-weighted probability matrix for the $i$-th instance is
computed as:%

\begin{equation}
\mathbf{P}^{(i)}=\mathbf{WS}^{(i)}\in \mathbb{R}^{M\times T}.
\label{class_prob_distr_matrix}%
\end{equation}

The output $\mathbf{P}^{(i)}$ provides refined event probabilities across $M$
generations and $T$ intervals for the $i$-th instance, where each element of
$\mathbf{P}^{(i)}$ is computed as a weighted sum of probabilities from
$\mathbf{S}^{(i)}$:
\begin{equation}
p_{k}^{(m)}(\mathbf{x}_{t})=\sum_{i=1}^{N}w_{k,i}\mathbf{S}_{m,t}^{(i)}.
\label{class_probability_1}%
\end{equation}

Here $w_{k,i}$ is an element of $\mathbf{W}$; $p_{k}^{(m)}(\mathbf{x}_{t})$ is
the $k$-th element of the class probability distribution $\mathbf{p}%
(\mathbf{x}_{t})$ for the $t$-th instance obtained through the $m$-th
generation of the probability distribution from the set $\mathcal{R}^{(t)}$;
$\mathbf{S}_{m,t}^{(i)}$ is a vector from the matrix $\mathbf{S}^{(i)}$
corresponding to the $m$-th generation from $\mathcal{R}^{(t)}$.

It is important to note that there are no generations for uncensored
observations because we have the precise probability distribution
(\ref{uncensor_distr}). Generations are performed only for censored
observations. In order to simplify the model description, we will assume that
$M$ generations are performed for all instances. However, the software
implementing the considered models distinguishes these cases.

\subsection{The first model}

The main idea behind the first proposed model, \emph{iSurvM}, is based on
averaging the loss function for learning the class probability distributions
$\mathbf{p}^{(m)}(\mathbf{x}_{i})$ over all generations with indices
$m=1,...,M$. The loss function based on using a likelihood function can be
written in the following form:
\begin{equation}
l(\mathbf{p}^{(m)}(\mathbf{x}_{i}))=-\mathbf{1}[\delta_{i}=1]\cdot \log \left(
p_{c(i)}^{(m)}(\mathbf{x}_{i})\right)  -\mathbf{1}[\delta_{i}=0]\cdot
\log \left(  \sum_{j=c(i)+1}^{T}p_{j}^{(m)}(\mathbf{x}_{i})\right)  ,
\label{Loss1}%
\end{equation}
where $p_{c(i)}^{(m)}(\mathbf{x}_{i})$ is the element of the class probability
distribution $\mathbf{p}^{(m)}(\mathbf{x}_{i})$ with the index $c(i)$.

The first term in the loss function (\ref{Loss1}) corresponds to uncensored
observations, the second term is responsible for censored observations. The
next step is averaging the loss functions $l(\mathbf{p}^{(m)}(\mathbf{x}%
_{i}))$ over all generations:
\begin{equation}
L(\mathbf{p}(\mathbf{x}_{i}))=\sum_{m=1}^{M}l(\mathbf{p}^{(m)}(\mathbf{x}%
_{i})). \label{Loss1_1}%
\end{equation}

Finally, we obtain the expected loss function for learning the class
probability distribution as follows:
\begin{equation}
\mathcal{L}(\mathbf{p})=\sum_{i=1}^{N}l(\mathbf{p}(\mathbf{x}_{i})).
\label{Loss2}%
\end{equation}

Learning the class probability distribution means that we learn attention
weights minimizing the loss function $\mathcal{L}(\mathbf{p})$ because
$\mathbf{p}(\mathbf{x})$ is the function of $\mathbf{\pi}_{k}$ (see
(\ref{class_prob_distr})). After finalizing the attention weights $\mathbf{W}%
$, we additionally learn the final class probability distribution. In this
learning process, the probabilities are initialized as averaged generated
values, and then they are optimized using the loss function (\ref{Loss1}).
This approach allows us to enhance the model accuracy.

An implementation of training the model iSurvM is presented as Algorithm
\ref{alg:iSurvM}.

\begin{algorithm}
\caption{The training algorithm for iSurvM} \label{alg:iSurvM}

\begin{algorithmic}
[1]\REQUIRE Training set $\mathcal{A}$; hyperparameters $p_{mask}$, $M$, $T$,
$d$; the number of epochs $E$

\ENSURE Weight matrix $\mathbf{W}$; probability distributions $\mathbf{\pi
}_{k},\ k=1,...,T$

\STATE Initialize randomly matrices $\mathbf{W}_{Q}$, $\mathbf{W}_{K}$ and
compute the attention matrix $\mathbf{A}$ using (\ref{Attention_1})

\STATE Compute the mask matrix $\mathbf{M}$ using (\ref{Attention_mask})

\STATE Compute the matrix $\mathbf{W}$ using (\ref{Weight_matrix_W})

\WHILE{Number of epochs $ \leq E$}

\STATE Generate probability distributions $\mathbf{q}_{1}^{(i)},...,\mathbf{q}%
_{M}^{(i)}$ from $\mathcal{R}^{(i)}$ for the $i$-th instance using the
Dirichlet distribution

\STATE Form the matrix $\mathbf{S}^{(i)}$, $i=1,...,N$

\STATE Compute the matrix $\mathbf{P}^{(i)}$ through $\mathbf{W}$ and
$\mathbf{S}^{(i)}$ using (\ref{class_prob_distr_matrix}), $i=1,...,N$

\STATE Learn $\mathbf{W}$ using the loss functions (\ref{Loss1})-(\ref{Loss2})

\STATE Compute the matrix $\mathbf{P}^{(i)}$ through $\mathbf{W}$ using
(\ref{class_prob_distr_matrix})

\ENDWHILE

\STATE Fine-tune the averaged values of the probability distributions
$\mathbf{q}_{1}^{(i)},...,\mathbf{q}_{M}^{(i)}$ generated at the last epoch
using the matrix $\mathbf{W}$ obtained also at the last epoch and using the
loss function (\ref{Loss2})
\end{algorithmic}
\end{algorithm}

To find the class probability distribution $(p_{1}(\mathbf{x}_{0}%
),...,p_{T}(\mathbf{x}_{0}))$ for a new instance $\mathbf{x}_{0}$ (the
inference phase) based on the trained neural network with attention parameters
$\mathbf{W}_{Q}$, $\mathbf{W}_{K}$, we use (\ref{class_prob_distr}) and the
trained matrix $\mathbf{W}$, but the vectors of probabilities $\mathbf{\pi
}_{1},...,\mathbf{\pi}_{T}$ in (\ref{class_prob_distr}) are replaced with
vectors $\mathbf{q}_{1}^{(i)},...,\mathbf{q}_{M}^{(i)}$ generated at the last
epoch in Algorithm \ref{alg:iSurvM}.

\subsection{The second model}

The main idea behind the second proposed model, \emph{iSurvQ}, is based on
averaging the same loss function (\ref{Loss1}) for learning the class
probability distributions $\mathbf{p}^{(m)}(\mathbf{x}_{i})$ over all
instances $i=1,...,N$, i.e.,
\begin{equation}
L(\mathbf{p}^{(m)}(\mathbf{x}))=\sum_{i=1}^{N}l(\mathbf{p}^{(m)}%
(\mathbf{x}_{i})). \label{Loss3}%
\end{equation}

Then, among all $L(\mathbf{p}^{(m)}(\mathbf{x}))$, $m=1,...,M$, a portion $r$
of the largest values of the loss functions are selected and averaged over the
selected generations. Here $r$ is a hyperparameter. Let $\mathcal{K}$ be the
set of indices corresponding to $r\cdot M$ largest values of $L(\mathbf{p}%
^{(m)}(\mathbf{x}))$. In fact, only $r\cdot M$ worse cases of the loss
function are taken into account. This implies that the model is robust.

The final expected loss function for learning the class probability
distribution is defined as follows:
\begin{equation}
\mathcal{L}(\mathbf{p})=\sum_{m\in \mathcal{K}}l(\mathbf{p}^{(m)}(\mathbf{x})).
\label{Loss4}%
\end{equation}

In this model, we again learn the attention weights $\mathbf{W}$, and then we
learn the final class probability distribution in the same way as it has been
described for iSurvM. Therefore, the algorithm implementing the model is
coincides with Algorithm \ref{alg:iSurvM} except for the step when the
attention weights $\mathbf{W}$ are trained. In iSurvQ, the matrix $\mathbf{W}$
is trained using the loss functions (\ref{Loss3}) and (\ref{Loss4}) instead of
(\ref{Loss1_1}) and (\ref{Loss2}).

\subsection{The third model}

The main difference of the third proposed model, \emph{iSurvJ}, is that the
class probabilities and the attention weights are jointly learned. This
learning allows us to avoid the generation procedure for the probability
distributions $\mathbf{q}_{1}^{(i)},...,\mathbf{q}_{M}^{(i)}$. To implement
the learning, the probability logits are initially initialized as zeros. At
each epoch, the $\mathrm{softmax}(\mathrm{logits})$ operation is applied to
the logits to obtain probabilities of intervals $\widetilde{\pi}_{k}^{(i)}$
and ensuring the condition $\sum_{k=1}^{T}\widetilde{\pi}_{k}^{(i)}=1$. Here
$\widetilde{\pi}_{k}^{(i)}$ are precise analogues of interval-valued
probabilities $\pi_{k}^{(i)}$ obtained by means of learning.

Let $p_{c(i)}(\mathbf{x}_{i})$ be the probability of the interval with the
index $c(i)$ defined using (\ref{N-W-2}), but the probability $\pi_{k}^{(i)}$
in (\ref{N-W-2}) is replaced with the precise probability $\widetilde{\pi}%
_{k}^{(i)}$. The loss function for learning the model is defined in
(\ref{Loss2}). However, the loss function $l(\mathbf{p}(\mathbf{x}_{i}))$ is
determined as
\begin{align}
l(\mathbf{p}(\mathbf{x}_{i}))  &  =-\mathbf{1}[\delta_{i}=1]\cdot \log \left(
p_{c(i)}(\mathbf{x}_{i})\right)  -\mathbf{1}[\delta_{i}=0]\cdot \log \left(
\sum_{j=c(i)+1}^{T}p_{j}(\mathbf{x}_{i})\right) \nonumber \\
&  -\gamma \sum_{k=1}^{T}\widetilde{\pi}_{k}^{(i)}\cdot \log \left(
\widetilde{\pi}_{k}^{(i)}\right)  . \label{Loss_regul}%
\end{align}

Here $\mathbf{p}(\mathbf{x}_{i})=(p_{1}(\mathbf{x}_{i}),...,p_{T}%
(\mathbf{x}_{i})\mathbf{)}$ is the precise probability distribution of the
time intervals for the $i$-th instance; $\gamma$ is the hyperparameter. It can
be seen from (\ref{Loss_regul}) that the regularization term is added to the
loss function, which corresponds to the entropy of the probability
distribution $\widetilde{\pi}^{(i)}$. The same term is added to the loss
functions (\ref{Loss1}) in iSurvM and iSurvQ when the last step in Algorithm
\ref{alg:iSurvM} is performed (fine-tuning the averaged values of the
probability distributions $\mathbf{q}_{1}^{(i)},...,\mathbf{q}_{M}^{(i)}$).
This regularization term prevents the model iSurvJ from making overly
optimistic decisions when optimizing both the weights and probabilities at the
same time.

An implementation of training the model iSurvM is presented as Algorithm
\ref{alg:iSurvM}.

\begin{algorithm}
\caption{The training algorithm for iSurvJ} \label{alg:iSurvJ}

\begin{algorithmic}
[1]\REQUIRE Training set $\mathcal{A}$; hyperparameters $p_{mask}$, $T$, $d$;
the number of epochs $E$

\ENSURE Weight matrix $\mathbf{W}$; probability distributions $\widetilde
{\mathbf{\pi}}_{k},\ k=1,...,T$

\STATE Initialize randomly matrices $\mathbf{W}_{Q}$, $\mathbf{W}_{K}$ and
compute the matrix $\mathbf{A}$ using (\ref{Attention_1})

\STATE Compute the mask matrix $\mathbf{M}$ using (\ref{Attention_mask})

\STATE Compute the matrix $\mathbf{W}$ using (\ref{Weight_matrix_W})

\STATE Initialize logits of $\widetilde{\pi}_{k}^{(i)}$ as zeros

\WHILE{Number of epochs $ \leq E$}

\STATE Compute $\widetilde{\pi}_{k}^{(i)}$ using the softmax operation applied
to logits $y$

\STATE Compute the probability distribution $\mathbf{p}(\mathbf{x}_{i})$ using
$\mathbf{W}$ and $\widetilde{\pi}_{k}^{(i)}$

\STATE Learn $\mathbf{W}$ and $\widetilde{\pi}_{k}^{(i)}$ using the loss
functions (\ref{Loss_regul}) and (\ref{Loss2})

\STATE Compute logits of $\widetilde{\pi}_{k}^{(i)}$

\ENDWHILE

\end{algorithmic}
\end{algorithm}

\subsection{Extended intervals}

Let us consider a dataset that contains only uncensored instances, where all
event times are unique. Recall that the attention mask $\mathbf{M}$ in
(\ref{Attention_mask}) is applied to the attention matrix to prevent examples
from attending to themselves. For uncensored examples (with $\delta=1$), a
constraint is imposed on the probabilities: the probability is set to $1$ in
the interval containing the actual event time, and $0$ in all other intervals.
This results in a probability vector of the form $[0,...,1,...,0]$. Since all
event times are unique (and hence the probability vectors as well), the only
example that could influence the learning of a given example is the example
itself. However, self-attention is removed by the mask, and thus the predicted
probability of falling into the correct interval, which is computed as a
weighted sum of probabilities, becomes 0 (since all the weighted terms are
masked out).

To fix this issue, instead of considering only the probability of the event in
the exact interval, we also consider the neighboring $2k$ intervals (looking
$k$ intervals on right and $k$ intervals on left), and sum their
probabilities. That is, the first term in the loss function, which was
\begin{equation}
-I[\delta_{i}=1]\cdot \log(S(t_{c(i)})-S(t_{c(i)+1})=-I[\delta_{i}=1]\cdot
\log(p_{c(i)}),
\end{equation}
is replaced with%
\begin{equation}
-I[\delta_{i}=1]\cdot \log(S(t_{c(i)-k})-S(t_{c(i)+1+k})=-I[\delta_{i}%
=1]\cdot \log \left(  \sum_{j=c(i)-k}^{c(i)+1+k}p_{j}\right)  ,
\label{loss_with_k}%
\end{equation}
where $k$ is a model hyperparameter that determines how many intervals around
the true event time are considered for uncensored data; $S(t_{c(i)})$ is the
SF which is defined through probabilities $p_{j}$.

As a result, a total of $2k+1$ intervals are used for analyzing each time
interval $\tau_{i}$.

This modification is applied to all the considered models and ensures that the
model considers not only the exact event interval but also adjacent intervals,
making training more robust and preventing zero gradients due to masking.

\section{Numerical experiments}

The codes implementing the proposed models are publicly available at: \url{https://github.com/NTAILab/iSurvMQJ}.

To rigorously evaluate the performance of the proposed models, we conducted
extensive experiments on 11 publicly available survival analysis datasets,
including \emph{Veterans, AIDS, Breast Cancer, WHAS500, GBSG2, BLCD, LND, GCD,
CML, Rossi, METABRIC}.

In addition, we study the models on synthetic dataset: \emph{Friedman1,2,3,
Linear, Quadratic, Strong Feature Interactions, Sparse Features, Nonlinear,
Noisy}. A complete description of the synthetic and real datasets can be found
in Appendix. Number of instance in the synthetic datasets for most
experiments: $500$ in the training set, $300$ in the test set. The noise
parameter in the datasets is $\epsilon=5$, except for the noisy data, where
$\epsilon=1$. The proportion of censored data in the datasets is $0.2$. Prior
to model training, all numerical features were standardized using z-score
normalization, while categorical features were encoded using one-hot encoding
to ensure compatibility with the machine learning algorithms.

The experimental design employed a nested cross-validation approach to ensure
robust performance estimation. The outer loop consists of four iterations of
5-fold stratified cross-validation, preserving the distribution of censored
events in each fold, with shuffling enabled and random seeds fixed for
reproducibility. Within each training fold of the outer loop, an additional
three-fold stratified cross-validation is performed for hyperparameter optimization.

We compared the performance of five models: the proposed models iSurvM,
iSurvQ, iSurvJ, and iSurvJ(G), along with the baseline Beran estimator. Here,
iSurvJ(G) is the same model as iSurvJ, but the attention weight is implemented
by using Gaussian kernel with the temperature parameter $\tau$ instead of the
neural network representation (\ref{Dimension1})-(\ref{Attention_1}) with
parameters $\mathbf{W}_{K}$, $\mathbf{W}_{Q}$, and $\mathbf{\theta}$.

Hyperparameter tuning was conducted using the Optuna library
\cite{akiba2019optuna}, which implements Bayesian optimization. Each model was
configured with a tailored search space encompassing key parameters such as
embedding dimensions (ranging from $64$ to $128$), learning rates (from
$10^{-4}$ to $1$), and regularization coefficients for weights (from $10^{-6}$
to $5\cdot10^{-2}$), and the entropy regularization (from $10^{-6}$ to $3$).
Additional architectural and training parameters included dropout rates ($0.3$
to $0.8$), mask rates ($0.1$ to $0.5$), batch sizes (10\% to 100\% of the
data), and epoch counts ($20$ to $2000$). The interval width parameter $k$,
which determines the granularity of predictions for uncensored examples, was
varied between $3$ and $10$ across different training instances.

Model performance was assessed using two well-established metrics in survival
analysis: the C-index, which measures the ranking consistency of predicted
risk scores, and the IBS, which evaluates the accuracy of probabilistic
predictions over time.

Results of numerical experiments with real and synthetic datasets illustrating
properties of the modes are presented in Appendix.

Table \ref{t:real-C-index} shows C-indices obtained for real datasets by using
the Beran estimator, iSurvM, iSurvQ, iSurvJ, and iSurvJ(G). It can be seen
from the presented results that the proposed models outperform the Beran
estimator almost for all datasets. If to compare the proposed models, then the
results imply that iSurvJ and iSurvJ(G) provide better results in comparison
with other models. Moreover, the worst results are observed for iSurvM. The
same conclusion can be made for real datasets based on the IBS as shown in
Table \ref{t:real-IBS}.

Numerical results for real data have shown that the best models are iSurvJ and
iSurvJ(G). Therefore, we study properties of these models on the basis of
synthetic data.

The proposed models consistently outperform the Beran estimator across
synthetic datasets (Linear, Quadratic, Strong Feature Interactions, etc.),
with the performance gap widening as the number of features increased (see
Figs. \ref{f:features_number_1_3}-\ref{f:features_4_6}). This suggests that
iSurvJ is better suited for datasets with complex, high-dimensional feature
structures. As the proportion of censored data increased, the Beran estimator
exhibits instability, while iSurvJ(G) maintains robust performance (see
Figs.~\ref{f:cens_isurvg_beran_1_3}-\ref{f:censor_isurvg_beran_7-9}). The
Kolmogorov-Smirnov (KS) distances between SFs further confirmed that
iSurvJ(G)'s predictions diverged significantly from the Beran estimator under
heavy censoring, highlighting its adaptability.

The interval-valued SFs generated by the interval-valued representation of the
instance probability distributions over the time intervals (see Figs.
\ref{f:intervals_friedman1} and \ref{f:intervals_nonlinear}) encapsulate the
Beran estimator's predictions within the bounds, suggesting that the proposed
model provides a more comprehensive representation of uncertainty, especially
for censored observations.

The parameter $k$, controlling the neighborhood of intervals for uncensored
data, initially improves accuracy but plateaues beyond a threshold (see
Fig.~\ref{f:k_ifriedman}). This indicates a trade-off between the
computational cost and marginal gains in predictive performance.

The unconditional SFs produced by iSurvJ closely matches those of the
nonparametric Kaplan-Meier estimator (see Figs.~\ref{f:uncond_veterans_gbsg2}
and~\ref{f:uncond_whas500_bc}), validating its consistency with classical
methods while offering conditional predictions.

To compare the performance of iSurvJ(G) and the Beran estimator, we present an
illustrative example using a synthetic dataset generated with a parabolic
time-to-event function (see Fig. \ref{f:parabola}). The example demonstrates
unstable behavior of the Beran estimator trained on specific complex data,
whereas iSurvJ(G) with the same kernel successfully captures the underlying
data structure.

In a nutshell, the experimental results demonstrate the superior performance
of the proposed iSurvJ and iSurvJ(G) models compared to the traditional Beran
estimator, particularly in high-dimensional and censored-data scenarios.

It should be noted that the models' performance relies on hyperparameter
tuning (e.g., embedding dimensions, regularization coefficients), though
Optuna-based optimization mitigated this. While iSurvJ(G) uses a simpler
Gaussian kernel, the neural network-based iSurvJ may face scalability issues
with large datasets.

\section{Conclusion}

This work has introduced three novel survival models: iSurvM, iSurvQ, and
iSurvJ, leveraging the imprecise probability theory and attention mechanisms
to address censored data challenges. One of the contributions includes
flexible modeling which means that the proposed framework avoids parametric
assumptions and accommodates interval-valued probabilities for censored
instances, enabling richer uncertainty quantification. We have also to point
out that iSurvJ and its Gaussian-kernel variant iSurvJ(G) consistently
outperformed the Beran estimator, particularly in high-dimensional settings
and under heavy censoring.

It should be noted that the proposed models deals with the standard survival
analysis task. At the same time, it is interesting to extend it to competing
risks and time-varying covariates. These are directions for further research.
We have considered only small and middle-dimensional data. Another important
direction for further research is to investigate scalability enhancements for
ultra-high-dimensional data.

The extreme case of imprecision when probabilities of time-to-event for the
censored observations are in intervals from 0 to 1. There exists imprecise
models and approaches which allows us to reduce the intervals employing
additional assumptions, for example, applying definitions of reachable
probability intervals \cite{Destercke-Antoine-2013} or survival models based
on imprecise Dirichlet distribution \cite{Coolen-97}.

\bibliographystyle{unsrt}
\bibliography{Attention,Boosting,Classif_bib,Convex,Deep_Forest,Explain,Imprbib,MYBIB,MYUSE,Surv_Attent,Survival_analysis}

\appendix

\section{Appendix: Experimental Details}

\subsection{Description of Synthetic Data}

To evaluate the performance of the survival analysis models, we generate
several synthetic datasets with varying levels of the complexity,
interactions, and noise. These datasets are designed to challenge models in
capturing non-linear dependencies, feature interactions, and the effects of censoring.

The censored data for all synthetic data are generated randomly according to
the Bernoulli distribution with some probabilities $\Pr \{ \delta_{i}=0\}$ and
$\Pr \{ \delta_{i}=1\}$. Event times are generated in accordance with the
Weibull distribution with the shape parameter $k>0$ and have the following form:%

\begin{equation}
T=\frac{y}{\Gamma \left(  1+\frac{1}{k}\right)  }\cdot \left(  -\log(u)\right)
^{\frac{1}{k}},
\end{equation}
where $u$ is a random number uniformly distributed in the interval $[0,1]$;
$y$ is a value that reflects a specific underlying relationship between
features and the event time, and is computed individually for each dataset as
it is shown below; $\Gamma(\cdot)$ is the gamma function.

The \emph{Friedman1}\textbf{ }dataset defines the variable $y$ as a non-linear
function of the input features $\mathbf{x}=(x_{1},x_{2},x_{3},x_{4},x_{5}%
)\in \mathbb{R}^{5}$, generated according to the following formula:
\begin{equation}
y=10\sin(\pi x_{1}x_{2})+20(x_{3}-0.5)^{2}+10x_{4}+5x_{5},
\end{equation}
where covariates $\mathbf{x}$ are independently drawn from the uniform
distribution on the interval $[0,1]$.

The remaining $d-5$ features (if $d>5$) are considered as noise variables and
do not influence the value of $y$.

The \emph{Friedman2}\textbf{ }dataset defines the variable $y$ through a
geometric relationship involving an ellipsoidal structure. The response is
computed as:
\begin{equation}
y=\sqrt{x_{1}^{2}+(x_{2}x_{3}-\frac{1}{x_{2}x_{4}})^{2}},
\end{equation}
where the input features $\mathbf{x}=(x_{1},x_{2},x_{3},x_{4})\in
\mathbb{R}^{4}$ are independently sampled from the uniform distribution on the
interval $[0,1]$. The features are then scaled and shifted as follows to
introduce variability and align with the desired characteristics of the
dataset: $x_{1}$ is scaled by the factor $100$; $x_{2}$ is scaled by $520\pi$
and shifted by $40\pi$; $x_{4}$ is scaled by $10$ and then shifted by $1$.

This function introduces strong non-linear interactions and is useful for
evaluating the performance of algorithms in capturing complex geometric
dependencies between features.

The \emph{Friedman3}\textbf{ }dataset defines the variable $y$ as a highly
non-linear function involving division, multiplicative interactions, and a
trigonometric transformation. Specifically, $y$ is computed as:
\begin{equation}
y=\arctan \left(  \frac{x_{2}x_{3}-1/(x_{2}x_{4})}{x_{1}}\right)  .
\end{equation}

As with the previous dataset, the features $\mathbf{x}=(x_{1},x_{2}%
,x_{3},x_{4})\in \mathbb{R}^{4}$ are generated by the same way.

This function introduces complex interactions between variables and a strong
non-linear transformation via the arctangent function, making it a challenging
benchmark for models.

To evaluate the models' capability to handle \emph{Strong feature
interactions}, we generate a dataset where the variable $y$ is determined as a
sum of the pairwise feature interactions:
\begin{equation}
y=\sum_{i=1}^{d}\sum_{j=i+1}^{d}w_{ij}x_{i}x_{j},
\end{equation}
where $\mathbf{x}=(x_{1},x_{2},\dots,x_{d})\in \mathbb{R}^{d}$ is a feature
vector, with each component independently drawn from the uniform distribution;
$\mathbf{W}=\{w_{ij}\} \in \mathbb{R}^{d\times d}$, $i,j=1,...,d$, is a
symmetric matrix of interaction coefficients with the zero diagonal, i.e.,
$w_{ij}=w_{ji}$, $w_{ii}=0,$ and for $i<j$, the upper-triangular elements are
independently sampled from a uniform distribution: $w_{ij}\sim \mathcal{U}%
(0,1)$.

This formulation ensures that the target variable $y$ is entirely governed by
second-order interactions.

The \emph{Sparse feature}\textbf{ }dataset represents a scenario with
high-dimensional sparse data. Features are generated as a sparse matrix
$X\in \mathbb{R}^{n\times d}$, where: $n$ is the number of samples, $d$ is the
number of features, the entries of $X$ are defined such that only some
proportion $s\in(0,1)$ of them are non-zero, with non-zero values typically
sampled from a uniform distribution:
\begin{equation}
x_{ij}\sim%
\begin{cases}
\mathcal{U}(0,1), & \text{with probability }s,\\
0, & \text{with probability }1-s.
\end{cases}
\end{equation}

The variable $\mathbf{y}\in \mathbb{R}^{n}$ is modeled as a linear function of
the features:
\begin{equation}
\mathbf{y}=X\mathbf{w},
\end{equation}
where $\mathbf{w}\in \mathbb{R}^{d}$ is a coefficient vector. The elements of
$\mathbf{w}$ are typically generated as:
\[
w_{j}\sim \mathcal{U}(0,1).
\]

This dataset design challenges models to handle sparsity in the feature space,
as most inputs contain limited information.

The \emph{Nonlinear dependency}\textbf{ }dataset is designed to model complex,
highly non-linear relationships between input features and the target
variable. The variable $y$ is defined as:%
\begin{equation}
y=4\sin(x_{1})+\log(|x_{2}|+1)+x_{3}^{2}+e^{0.5x_{4}}+\tanh(x_{5})+\sum
_{j=6}^{d}f_{j}(x_{j}),
\end{equation}
where $\mathbf{x}=(x_{1},x_{2},\dots,x_{d})\in \mathbb{R}^{d}$ is a feature
vector with $d\geq5$; each component $x_{j}\sim \mathcal{U}(0,1)$ sampled
independently; $f_{j}(x_{j})$ are additional nonlinear functions applied to
the remaining features $j=6,\dots,d$, which may be chosen from a predefined
set of nonlinear transformations (e.g., $\sin(x_{j})\cdot \sqrt{|x_{j-1}|+1}$,
$\log(|x_{j}|+1)\cdot \tanh(x_{j-2})$, $x_{j}^{2}\cdot \cos(x_{j-3})$) to
increase the overall complexity of the target variable.

This construction creates a highly non-linear mapping from features to the
target, requiring models to capture a variety of nonlinear behaviors including
oscillatory, exponential, polynomial, and saturation effects.

In the \emph{Noisy}\textbf{ }dataset, the target variable $T$ is generated by
applying a transformation $y=X\mathbf{w}$ using the Weibull distribution with
a small shape parameter $k\leq1$, leading to the high-variance noise. Here
$X\in \mathbb{R}^{n\times d}$ is a feature matrix with entries $x_{ij}%
\sim \mathcal{U}(0,1)$; $\mathbf{w}\in \mathbb{R}^{d}$ is a coefficient vector.
Elements of $\mathbf{w}$ are typically generated as: $w_{j}\sim \mathcal{U}%
(0,1)$.

By setting $k\leq1$, the resulting distribution of $T$ becomes heavy-tailed
and highly variable, which significantly obscures the underlying linear
relationship encoded by $y$, thus testing a model's robustness to noise.

The \emph{Linear}\textbf{ }dataset is designed to model simple linear
dependencies between features and the target variable. The variable $y$ is
generated as follows:%
\begin{equation}
y=\sum_{i=1}^{d}w_{i}x_{i}=\mathbf{x}^{\top}\mathbf{w},
\end{equation}
where: $\mathbf{x}=(x_{1},x_{2},\ldots,x_{d})^{\top}\in \mathbb{R}^{d}$ is a
feature vector with components sampled independently from the uniform
distribution $x_{i}\sim \mathcal{U}(0,1)$: $\mathbf{w}=(w_{1},w_{2}%
,\ldots,w_{d})^{\top}\in \mathbb{R}^{d}$ is a vector of weights with
coefficients also sampled independently from a uniform distribution $w_{i}%
\sim \mathcal{U}(0,1)$.

This setup ensures that the target variable $y$ is a pure linear function of
the input features, serving as a baseline for evaluating a model's ability to
learn linear relationships.

The \emph{Quadratic} dataset is designed to capture second-order (quadratic)
dependencies between features. The variable is constructed as a quadratic form
of the input features:%
\begin{equation}
y=\mathbf{x}^{\top}Q\mathbf{x},
\end{equation}
where $\mathbf{x}=(x_{1},x_{2},\ldots,x_{d})^{\top}\in \mathbb{R}^{d}$ is a
feature vector with components independently sampled from the uniform
distribution: $x_{i}\sim \mathcal{U}(0,1)$, $Q\in \mathbb{R}^{d\times d}$ is a
symmetric positive semi-definite matrix that defines the curvature of the
quadratic form $Q=A^{\top}A$; $A\in \mathbb{R}^{d\times d}$ is a matrix with
entries sampled from the standard normal distribution $A_{ij}\sim
\mathcal{N}(0,1)$.

This formulation ensures that $Q$ is symmetric and positive semi-definite,
thus guaranteeing $y\geq0$ for all $\mathbf{x}$, and that the target variable
captures interactions between features in a smooth, convex manner.

\subsection{Description of Real Data}

The \emph{Veterans' Administration Lung Cancer Study (Veteran) Dataset}
includes data on 137 patients characterized by 6 features. It is available
through the R package \textquotedblleft survival\textquotedblright.

\emph{The AIDS Clinical Trials Group Study (AIDS) Dataset} contains healthcare
and categorical data for 2139 patients, described by 23 features, all
diagnosed with AIDS. It can be accessed on Kaggle: \url{https://www.kaggle.com/datasets/tanshihjen/aids-clinical-trials}.

The \emph{(Breast Cancer) Dataset} consists of 198 samples described by 80
features. The endpoint is the occurrence of distant metastases, which are
observed in 51 patients (25.8\%). It can be obtained from the Python library
\textquotedblleft scikit-survival\textquotedblright.

The \emph{Worcester Heart Attack Study (WHAS500) Dataset} considers 500
patients with 14 features. The dataset can be obtained via the
\textquotedblleft smoothHR\textquotedblright \ R package or the Python
\textquotedblleft scikit-survival\textquotedblright \ package.

The \emph{German Breast Cancer Study Group 2 (GBSG2) Dataset} includes
observations of 686 patients with 10 features. It is accessible via the
\textquotedblleft TH.data\textquotedblright \ R package.

The \emph{Bladder Cancer Dataset (BLCD) } includes observations of 86 patients
after surgery assigned to placebo or chemotherapy. It has two features and can
be obtained from \url{https://www.stat.rice.edu/~sneeley/STAT553/Datasets/survivaldata.txt}.

The \emph{Lupus Nephritis Dataset (LND)} consists of observations of 87
persons with lupus nephritis, followed for over 15 years after an initial
renal biopsy. This dataset only contains time to death/censoring, indicator,
duration and log(1+duration), where duration is the duration of untreated
disease prior to biopsy. The dataset is available at \url{http://www.stat.rice.edu/~sneeley/STAT553/Datasets}.

The \emph{Gastric Cancer Dataset (GCD)} includes observations of 90 patients with
4 features. It is available through the R package \textquotedblleft
coxphw\textquotedblright .

The \emph{Chronic Myelogenous Leukemia Survival (CML) Dataset} is simulated
according to the structure of data by the German CML Study Group. The dataset
consists of 507 observations with 7 features: a factor with 54 levels
indicating the study center; a factor with levels trt1, trt2, trt3 indicating
the treatment group; sex (0 = female, 1 = male); age in years; risk group (0 =
low, 1 = medium, 2 = high); censoring status (FALSE = censored, TRUE = dead);
time survival or censoring time in days. The dataset can be obtained via the
\textquotedblleft multcomp\textquotedblright \ R package (cml).

The \emph{(Rossi) Dataset } contains 432 convicts released from Maryland state
prisons in the 1970s, described by 62 features. The dataset can be obtained
via the \textquotedblleft RcmdrPlugin.survival\textquotedblright \ R package.

The \emph{Molecular Taxonomy of Breast Cancer International Consortium
(METABRIC) Database} contains records of 1904 breast cancer patients,
including 9 gene indicators and 5 clinical features. The dataset is available
at \url{https://www.kaggle.com/datasets}.

\subsection{Study of the Model Properties}%

\begin{table}[tbp] \centering
\caption{C-indices obtained for real datasets by using the Beran estimator, iSurvM, iSurvQ, iSurvJ, and iSurvJ(G)}%
\begin{tabular}
[c]{lccccc}\hline
Dataset & Beran & iSurvM & iSurvQ & iSurvJ & iSurvJ(G)\\ \hline
Veterans & $0.7040$ & $0.7103$ & $\mathbf{0.7295}$ & $\underline{0.7196}$ &
$0.7003$\\
AIDS & $\mathbf{0.7529}$ & $0.7319$ & $0.7359$ & $0.7139$ & $\underline
{0.7483}$\\
Breast Cancer & $\underline{0.6519}$ & $0.6344$ & $0.6240$ & $0.6487$ &
$\mathbf{0.6600}$\\
WHAS500 & $0.7568$ & $\mathbf{0.7632}$ & $0.7596$ & $\underline{0.7616}$ &
$0.7575$\\
GBSG2 & $0.6730$ & $0.6827$ & $\mathbf{0.6883}$ & $\underline{0.6863}$ &
$0.6706$\\
BLCD & $0.5059$ & $\underline{0.5068}$ & $0.4835$ & $0.5067$ &
$\mathbf{0.5080}$\\
LND & $0.4695$ & $0.5264$ & $\underline{0.5663}$ & $\mathbf{0.5730}$ &
$0.4229$\\
GCD & $0.4535$ & $0.5201$ & $\mathbf{0.5717}$ & $\underline{0.5690}$ &
$0.4266$\\
CML & $0.6410$ & $0.6633$ & $\underline{0.6806}$ & $\mathbf{0.6917}$ &
$0.6424$\\
Rossi & $0.5817$ & $\underline{0.6088}$ & $0.6068$ & $0.5853$ &
$\mathbf{0.6186}$\\
METABRIC & $0.6261$ & $0.6410$ & $\underline{0.6418}$ & $\mathbf{0.6470}$ &
$0.6407$\\ \hline
\end{tabular}
\label{t:real-C-index}%
\end{table}%
%

\begin{table}[tbp] \centering
\caption{Brier score obtained for real datasets by using the Beran estimator, iSurvM, iSurvQ, iSurvJ, and iSurvJ(G)}%
\begin{tabular}
[c]{lccccc}\hline
Dataset & Beran & iSurvM & iSurvQ & iSurvJ & iSurvJ(G)\\ \hline
Veterans & $0.1373$ & $0.1189$ & $\mathbf{0.1221}$ & $\underline{0.1229}$ &
$0.1365$\\
AIDS & $\underline{0.0716}$ & $0.0791$ & $0.0805$ & $0.0786$ &
$\mathbf{0.0678}$\\
Breast Cancer & $\mathbf{0.1853}$ & $0.2539$ & $0.2377$ & $0.2425$ &
$\underline{0.2071}$\\
WHAS500 & $0.2174$ & $0.1896$ & $\underline{0.1900}$ & $\mathbf{0.1883}$ &
$0.2186$\\
GBSG2 & $0.2074$ & $0.2019$ & $\mathbf{0.1995}$ & $\underline{0.1998}$ &
$0.2186$\\
BLCD & $0.2952$ & $0.2701$ & $\underline{0.2783}$ & $0.2800$ &
$\mathbf{0.2716}$\\
LND & $\mathbf{0.2123}$ & $0.2332$ & $0.2194$ & $0.2333$ & $\underline
{0.2180}$\\
GCD & $0.1996$ & $0.1911$ & $\underline{0.1860}$ & $\mathbf{0.1850}$ &
$0.1991$\\
CML & $0.1456$ & $0.1325$ & $\underline{0.1372}$ & $\mathbf{0.1339}$ &
$0.1488$\\
Rossi & $0.3086$ & $\mathbf{0.1068}$ & $\underline{0.1083}$ & $0.1090$ &
$0.1084$\\
METABRIC & $0.2015$ & $0.2002$ & $\underline{0.1980}$ & $\mathbf{0.1952}$ &
$0.2078$\\ \hline
\end{tabular}
\label{t:real-IBS}%
\end{table}%

\subsubsection{Dependence on the number of features}

Using synthetic datasets, we can study how the number of features impacts the
accuracy measures. We apply the datasets: Linear, Quadratic, Strong Feature
Interactions, Sparse Features, Nonlinear, Noisy. The number of features ranges
from $1$ to $10$ in increments of $1$. The features are standardized.
Parameters of iSurvJ are the following: the number of epochs is $300$, the
learning rate is $10^{-2}$, $\gamma=0.1$, the mask rate is $0.5$, the
regularization coefficient for weights is $2\cdot10^{-3}$, the embedding
dimension is $64$, the batch\_rate is $0.2$, the dropout rate is $0.5$.
Parameters of the Beran estimator: the Gaussian kernel with the temperature
parameter $\tau=0.1$. For each number of features in one dataset, $50$
experiments were conducted.

\begin{figure}
[ptb]
\begin{center}
\includegraphics[
height=6.197in,
width=5.6542in
]%
{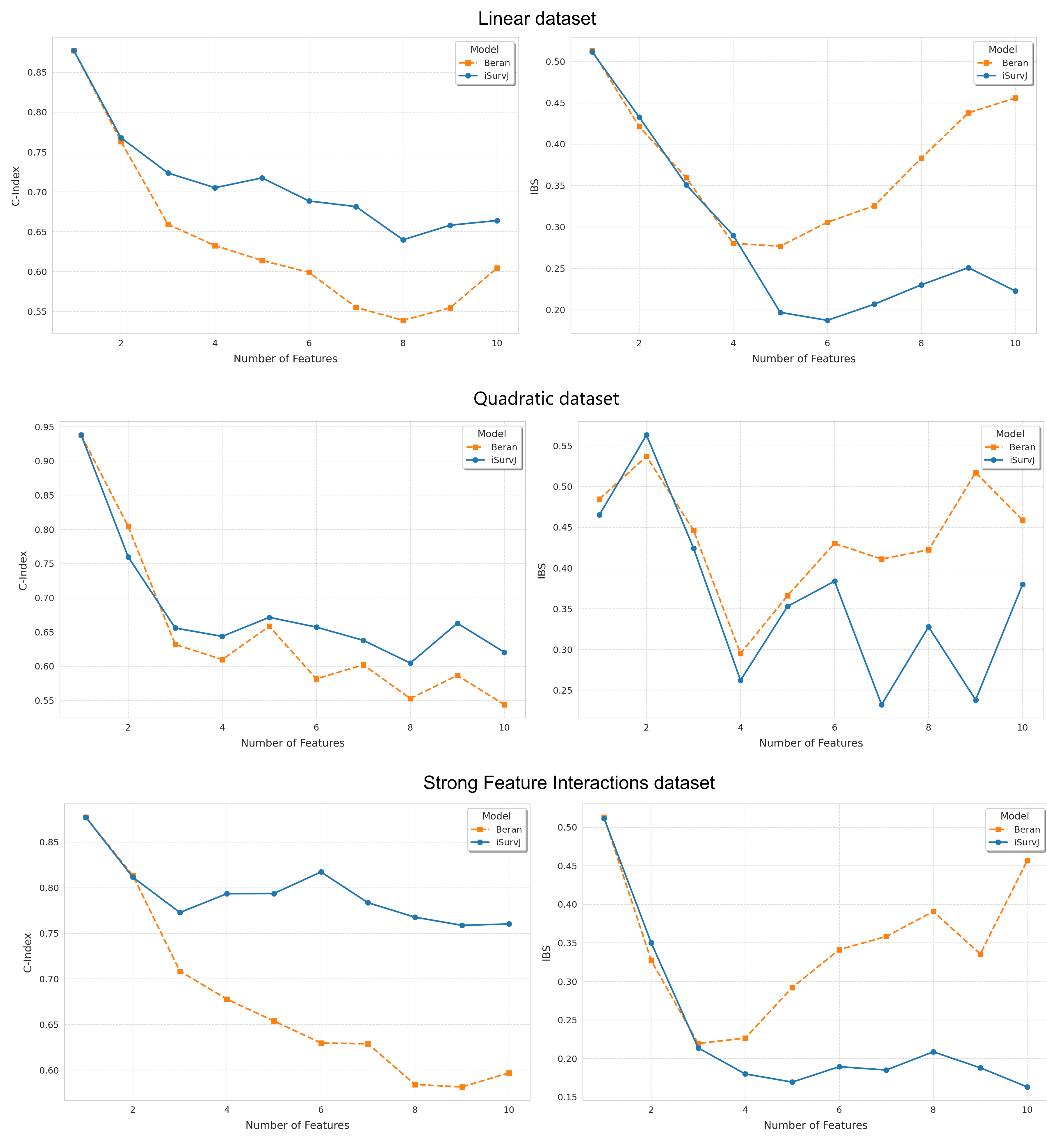}%
\caption{Dependence of the C-index and IBS on the number of features for the
Linear, Quadratic, Strong Feature Interactions datasets}%
\label{f:features_number_1_3}%
\end{center}
\end{figure}

\begin{figure}
[ptb]
\begin{center}
\includegraphics[
height=6.0117in,
width=5.6007in
]%
{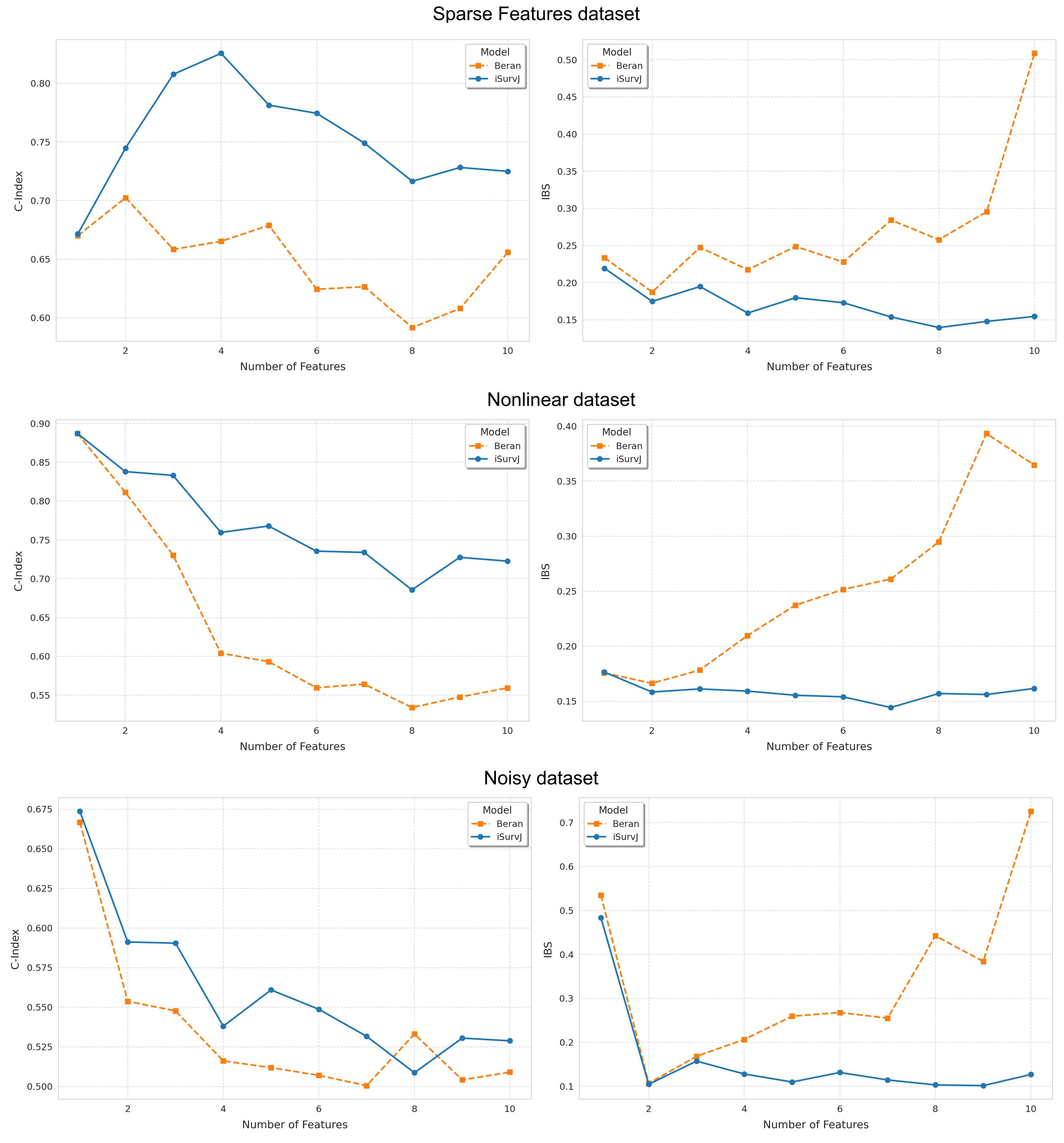}%
\caption{Dependence of the C-index and IBS on the number of features for the
Sparse Features, Nonlinear, Noisy datasets}%
\label{f:features_4_6}%
\end{center}
\end{figure}

Figs. \ref{f:features_number_1_3}-\ref{f:features_4_6}~show the dependence of
the C-index and IBS on the number of features. It can be seen from Figs.
\ref{f:features_number_1_3}-\ref{f:features_4_6} that iSurvJ demonstrates
better results compared to the Beran estimator all considered synthetic
datasets. Moreover, the greater the number of features in the instances, the
more significant the difference between the proposed model and the Beran
estimator. This implies that the proposed model is advisable to use for
datasets with high-dimensional feature vectors.%

%

\subsubsection{Dependence on the Parameter $k$}

The next question is how the parameter $k$ in (\ref{loss_with_k}) impact the
model accuracy. We apply the synthetic datasets: Friedman1, Friedman2,
Friedman3. Parameters of the model iSurvJ are the same as in the previous
experiments. The number of features is $5$. Proportion of censored data in the
dataset is $0.2$. Values of the parameter $k$ vary from 0 to 20 in increments
of 1.

\begin{figure}
[ptb]
\begin{center}
\includegraphics[
height=5.1048in,
width=5.5504in
]%
{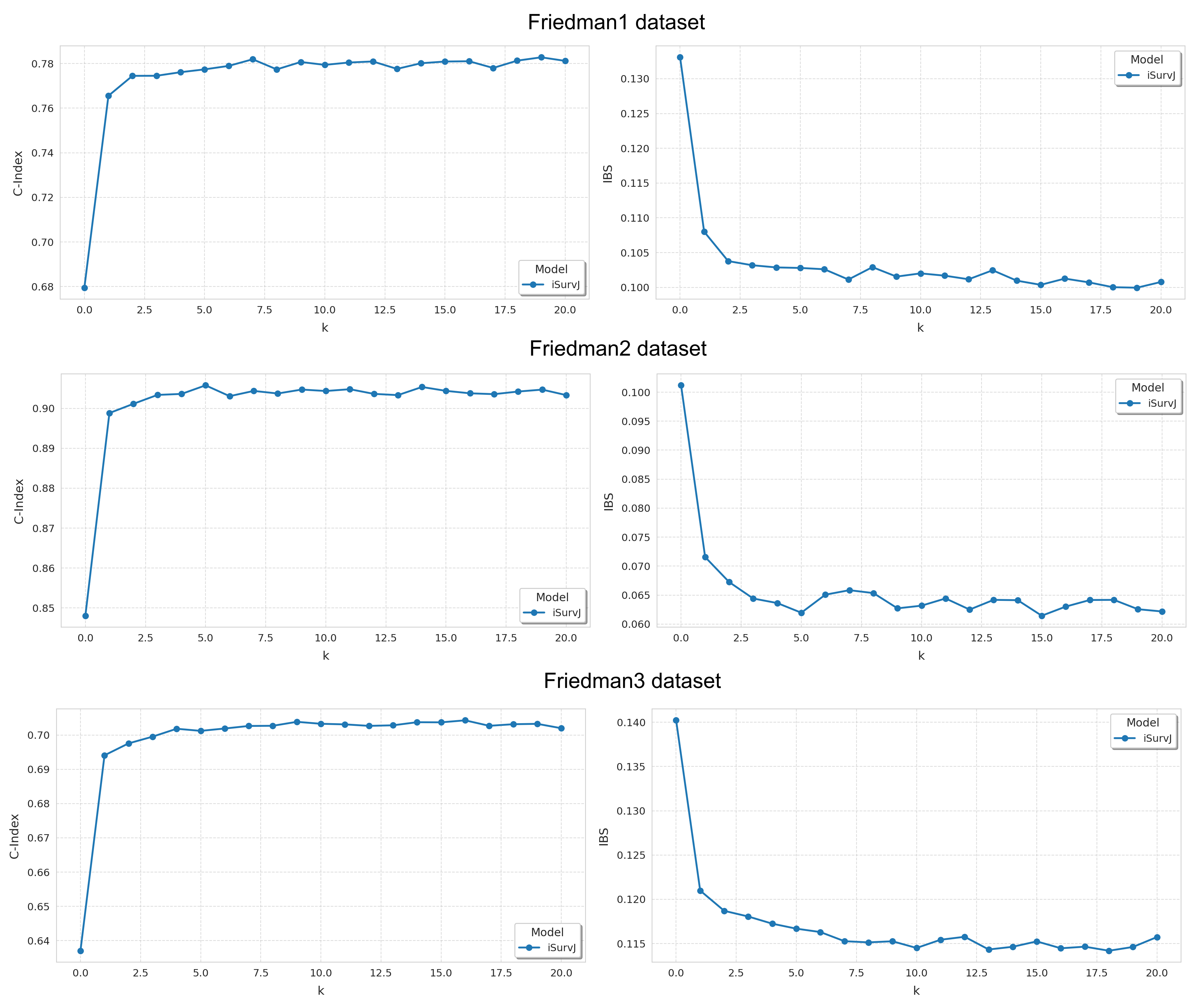}%
\caption{Dependence of the C-index and IBS on the parameter $k$ for the
Friedman1, Friedman2, Friedman3 datasets}%
\label{f:k_ifriedman}%
\end{center}
\end{figure}

Results are shown in Fig.~\ref{f:k_ifriedman}. It can be seen from
Fig.~\ref{f:k_ifriedman} that the model's accuracy significantly improves as
the parameter $k$ increases, but beyond a certain threshold, the improvement
plateaus. At the same time, increasing $k$ also raises the computational
complexity. Therefore, beyond a certain value, further increasing $k$ becomes pointless.%

\subsubsection{Dependence on the Number of Censored Observations}

Another question for studying is how the model accuracy depends on the number
of censored instances. We apply the synthetic datasets: Strong Feature
Interactions and Sparse Features with the number of features $5$. The
proportion of censored data in each dataset varies from $0$ to $0.8$ in
increments of $0.1$. Parameters of iSurvJ are the same as in the previous experiments.

\begin{figure}
[ptb]
\begin{center}
\includegraphics[
height=4.0539in,
width=5.7695in
]%
{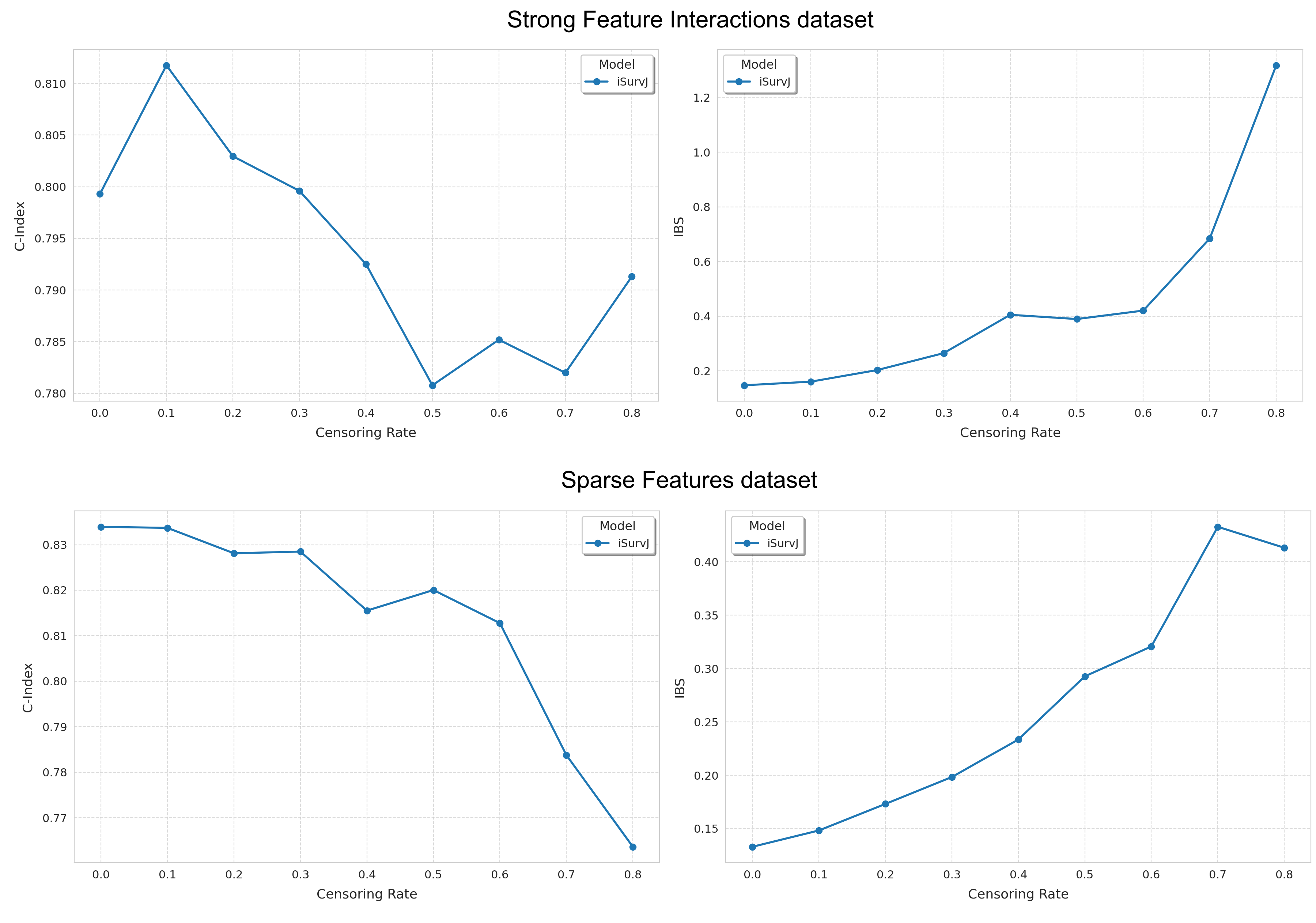}%
\caption{Dependence of the C-index and IBS on the proportion of censored data
for the Strong Feature Interactions and Sparse Features datasets}%
\label{f:isurvj_censor}%
\end{center}
\end{figure}

Fig.~\ref{f:isurvj_censor} shows the dependence of the C-index and IBS on the
proportion of censored data. The experimental results primarily illustrate how
the accuracy of the proposed model decreases as the proportion of censored
data increases.%

\begin{figure}
[ptb]
\begin{center}
\includegraphics[
height=5.0661in,
width=5.7563in
]%
{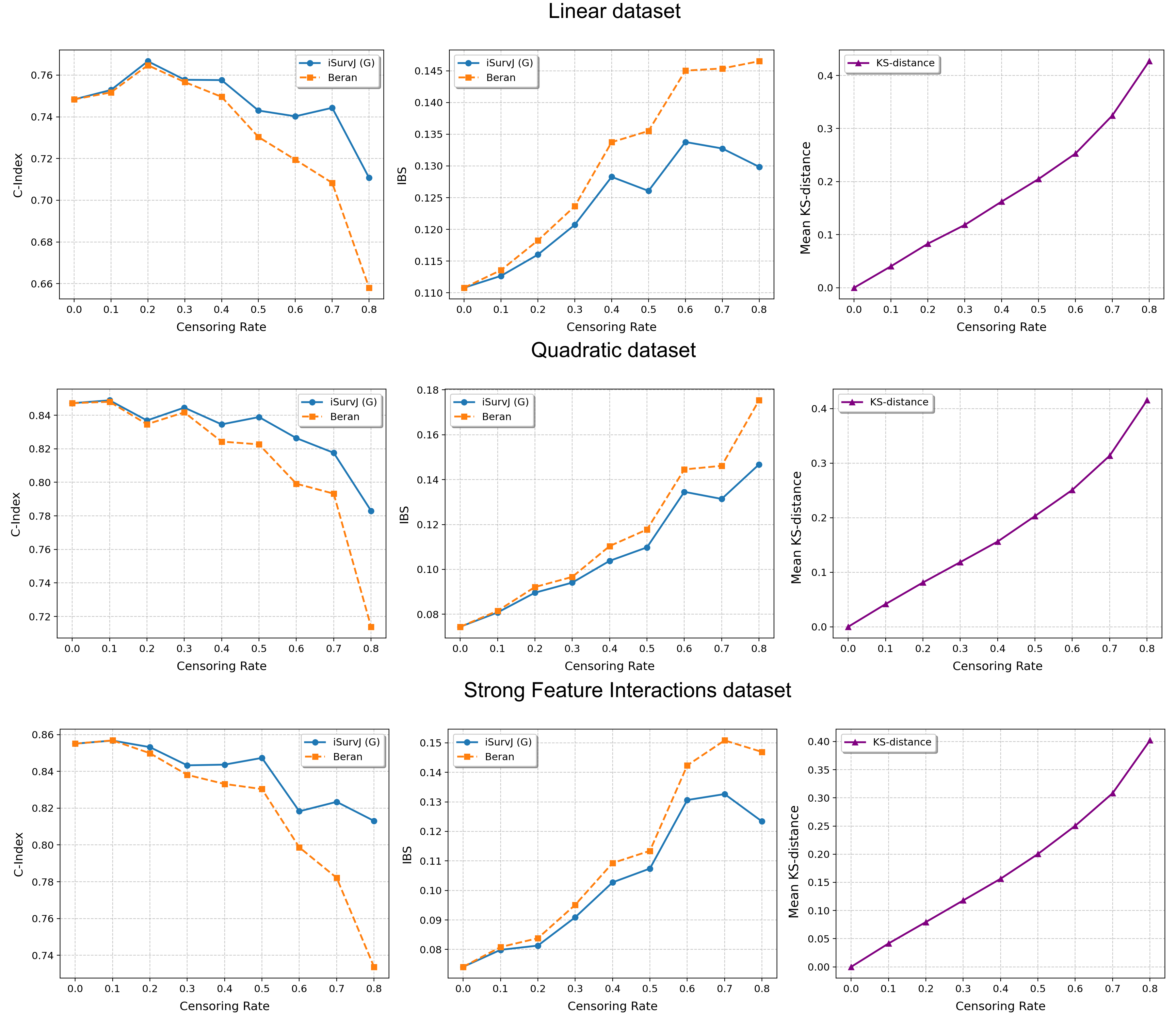}%
\caption{Comparison of iSurvJ(G) and the Beran estimator by different values
of the censored observation proportion for Linear, Quadratic, Strong Feature
Interactions datasets}%
\label{f:cens_isurvg_beran_1_3}%
\end{center}
\end{figure}

\begin{figure}
[ptb]
\begin{center}
\includegraphics[
height=5.2613in,
width=6.0084in
]%
{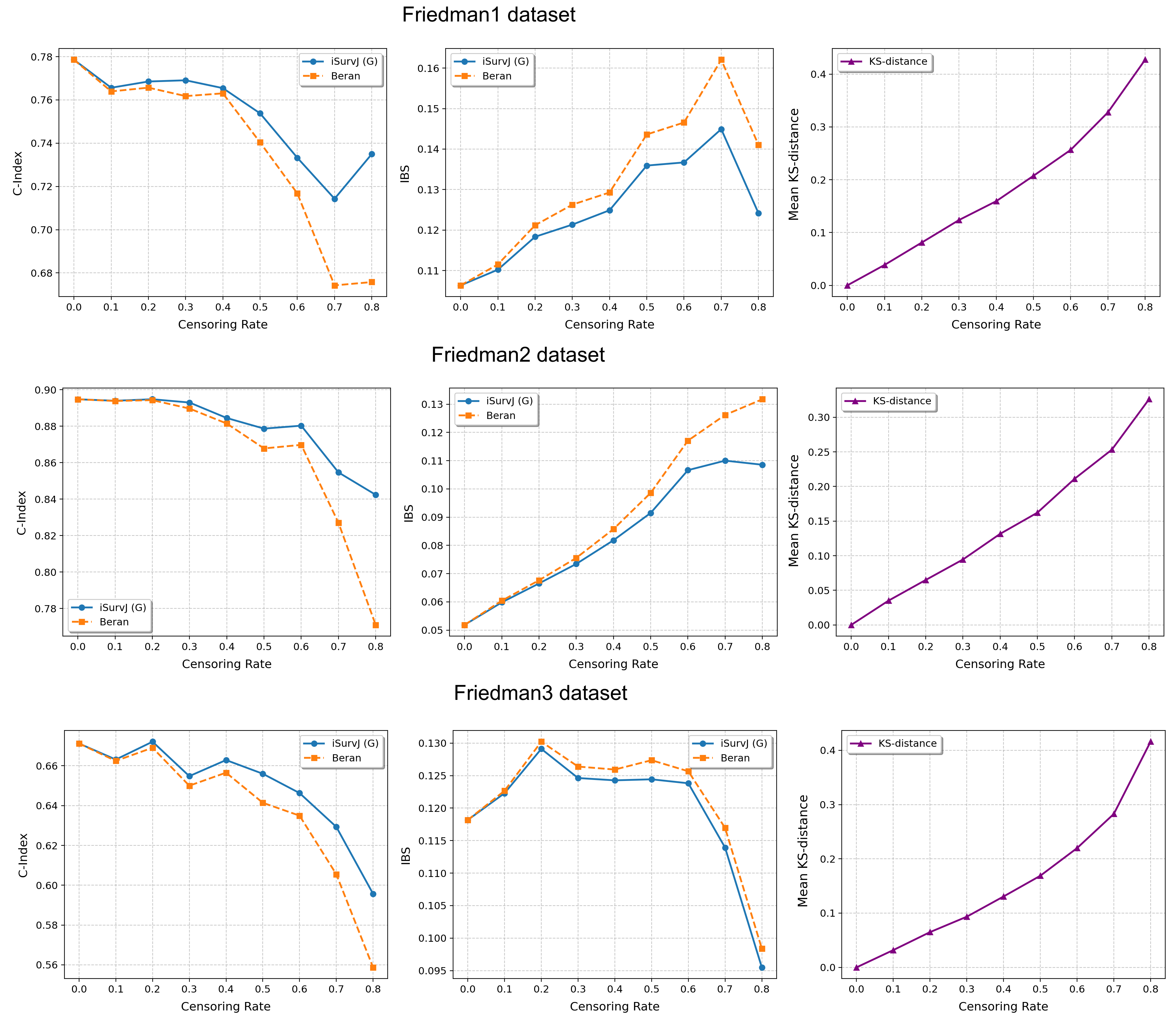}%
\caption{Comparison of iSurvJ(G) and the Beran estimator by different values
of the censored observation proportion for the Friedman1, Friedman2, Friedman3
datasets}%
\label{f:cens_isurvg_beran_4_6}%
\end{center}
\end{figure}

\begin{figure}
[ptb]
\begin{center}
\includegraphics[
height=5.2465in,
width=5.8544in
]%
{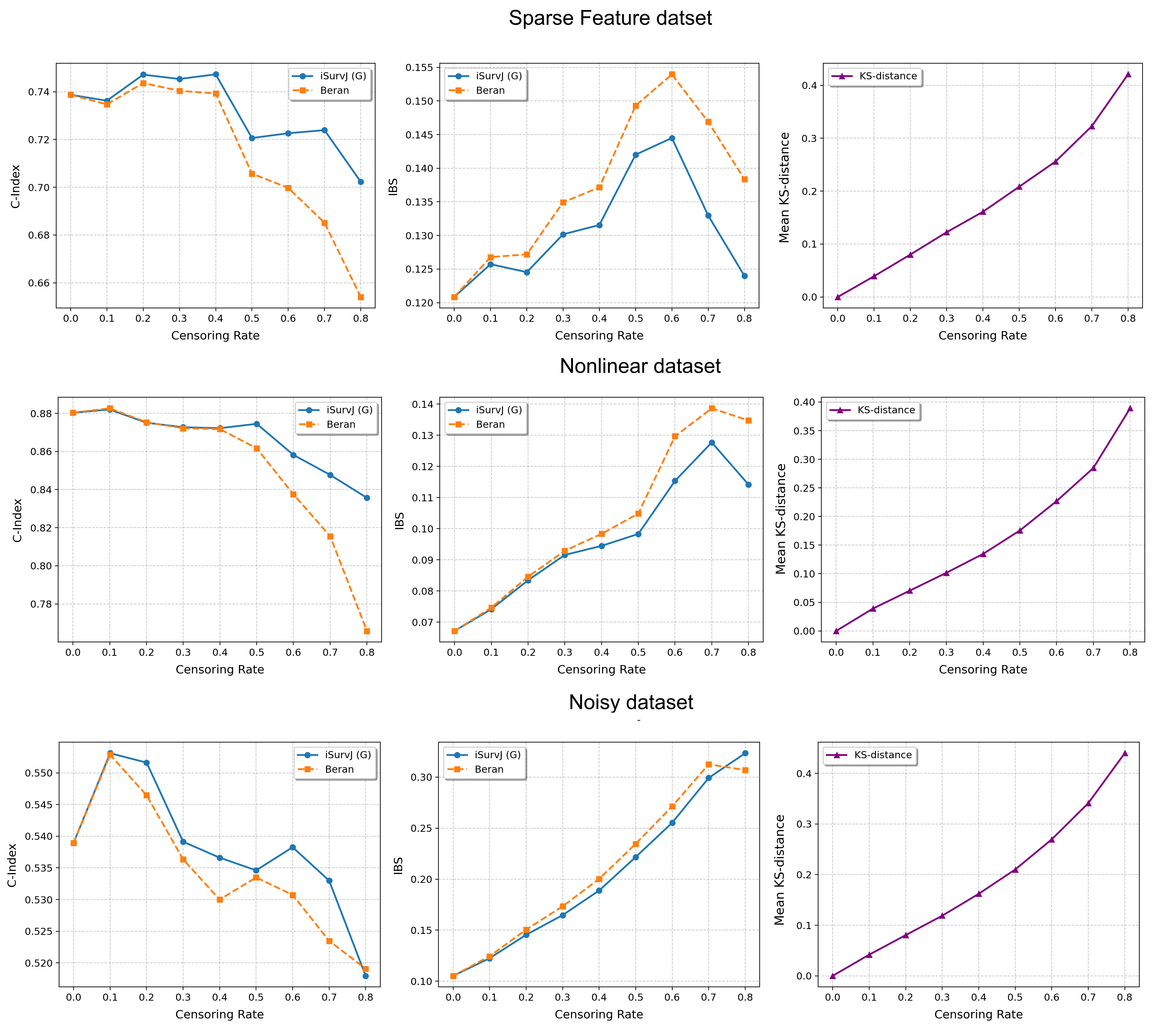}%
\caption{Comparison of iSurvJ(G) and the Beran estimator by different values
of the censored observation proportion for the Sparse Features, Nonlinear,
Noisy datasets}%
\label{f:censor_isurvg_beran_7-9}%
\end{center}
\end{figure}

The comparative experiments are illustrated in
Figs.~\ref{f:cens_isurvg_beran_1_3}-\ref{f:censor_isurvg_beran_7-9}, which
show the proposed model compared to the Beran estimator as the proportion of
censored data increases. The average Kolmogorov-Smirnov (KS) distances between
SFs obtained by these models for the same instances are also shown, depending
on the proportion of censored data. The comparison results are provided for
datasets: Linear, Quadratic, Strong Feature Interactions, Friedman1,
Friedman2, Friedman3, Sparse Features, Nonlinear, Noisy. The key feature of
these comparative experiments is that instead of the iSurvJ model, we use the
iSurvJ(G) model, where the neural network attention mechanism is replaced with
a Gaussian kernel featuring a learnable parameter. It is interesting to point
out from Figs.~\ref{f:cens_isurvg_beran_1_3}-\ref{f:censor_isurvg_beran_7-9}
that iSurvJ(G) outperforms the Beran estimator especially when the number of
censored instances is large. This is clearly demonstrated by the KS distances
which show how the difference between SFs of two models increases with the
proportion of censored data.%

%

%

\subsubsection{Intervals for Survival Functions}

\begin{figure}
[ptb]
\begin{center}
\includegraphics[
height=4.5in,
width=5.9in
]%
{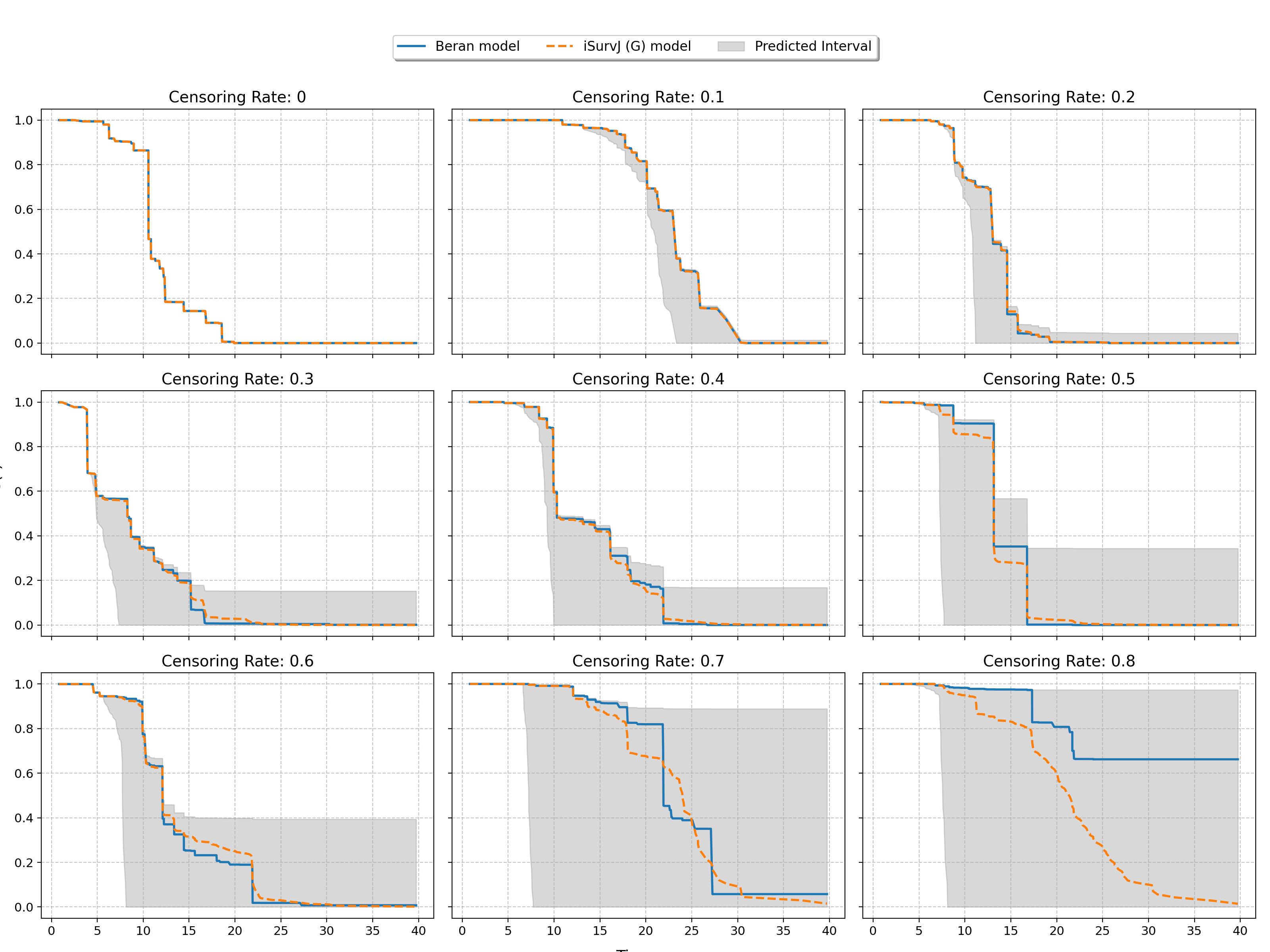}%
\caption{Interval-valued SFs for the Friedman1 dataset by different values of
the censiring rate}%
\label{f:intervals_friedman1}%
\end{center}
\end{figure}

\begin{figure}
[ptb]
\begin{center}
\includegraphics[
height=4.5in,
width=5.9in
]%
{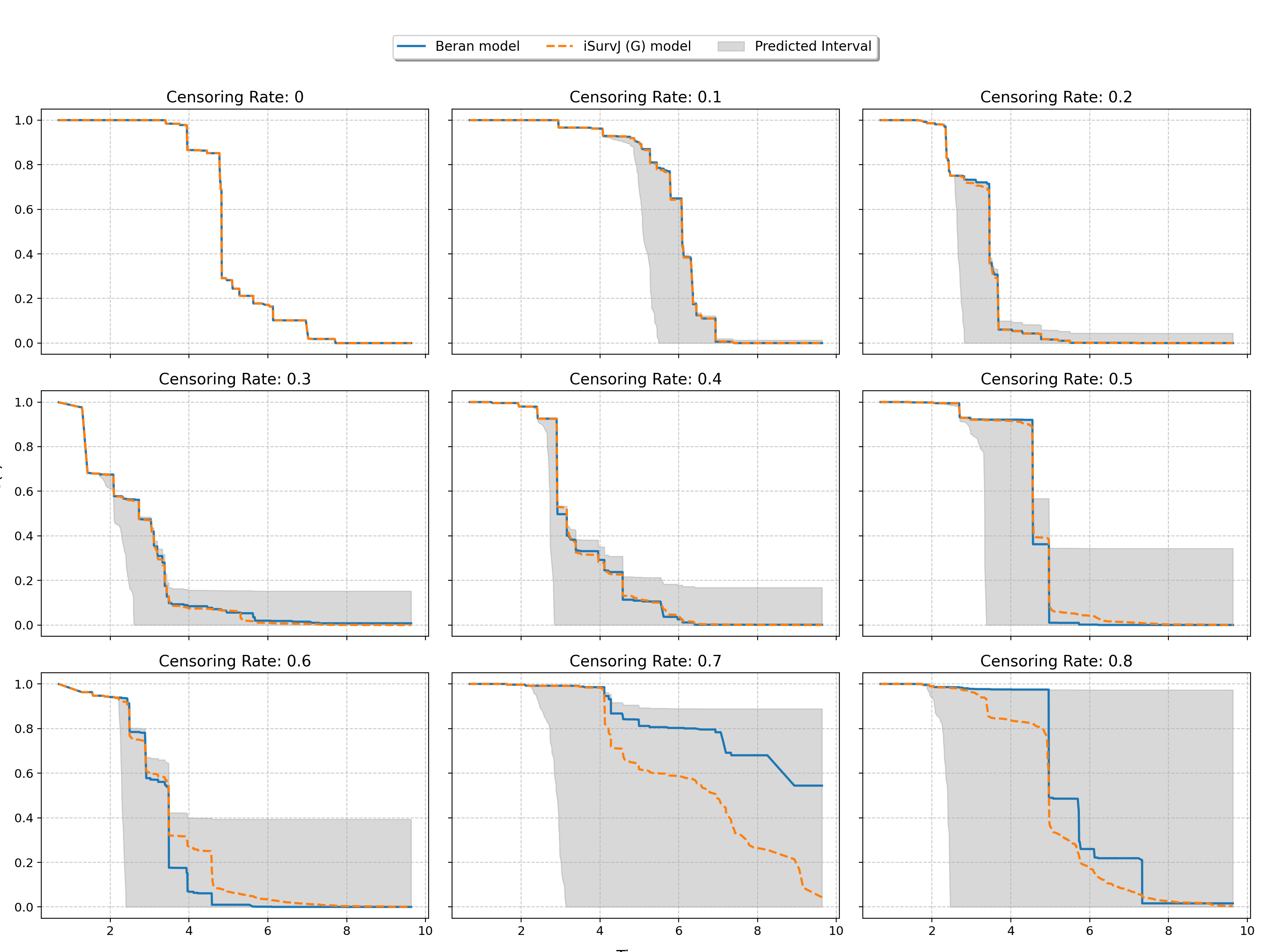}%
\caption{Interval-valued SFs for the Nonlinear dataset by different values of
the censiring rate}%
\label{f:intervals_nonlinear}%
\end{center}
\end{figure}

The next experiments illustrate intervals of SFs depicted in Figs.
\ref{f:intervals_friedman1} and~\ref{f:intervals_nonlinear} are constructed as
follows: the plots are shown for one instance from a dataset (similar patterns
can be observed for other instances and other datasets). One SF is produced as
the prediction for this instance by the iSurvJ(G) model. Another SF is
obtained as a prediction using the Beran estimator.

The bounds for the SF are obtained using (\ref{N-W-2}), where probabilities
$\pi_{k}^{(i)}$, $k=1,...,T$, are interval-valued for censored observations.
The attention weights $a_{j,i}(\mathbf{w})$ are computer by using the Gaussian
kernels. As a result, we get the interval-valued probability distribution
$p_{k}(\mathbf{x}_{j})$, $k=1,...,T$, for the $j$-th instance, which produces
the interval-valued SFs, depicted in Figs. \ref{f:intervals_friedman1} and
\ref{f:intervals_nonlinear} for instances from the Friedman1 and Nonlinear
datasets, respectively.

It is important to note that the Beran estimator predicts the SF which is
totally inside the lower and upper SF bounds. This is a very interesting observation.%

%

\subsubsection{Unconditional Survival Functions}

Let us study the relationship between the unconditional SFs produced by iSurvJ
and the Kaplan-Meier estimator. The unconditional SF is calculated by
averaging all conditional SFs obtained for all instances from a dataset. We
consider real datasets: Veterans Lung Cancer, GBSG2, WHAS500, Breast Cancer.
Numerical features are standardized, categorical features are encoded.
Parameters of iSurvJ are the following: the number of epochs is $1000$, the
learning rate is $10^{-2}$, $\gamma=0.1$, the mask rate is $0$, the
regularization coefficient for weights is $2\cdot10^{-3}$, the embedding
dimension is $64$, the batch\_rate is $0.1$, the dropout rate is $0.6$.

\begin{figure}
[ptb]
\begin{center}
\includegraphics[
height=5.0925in,
width=4.161in
]%
{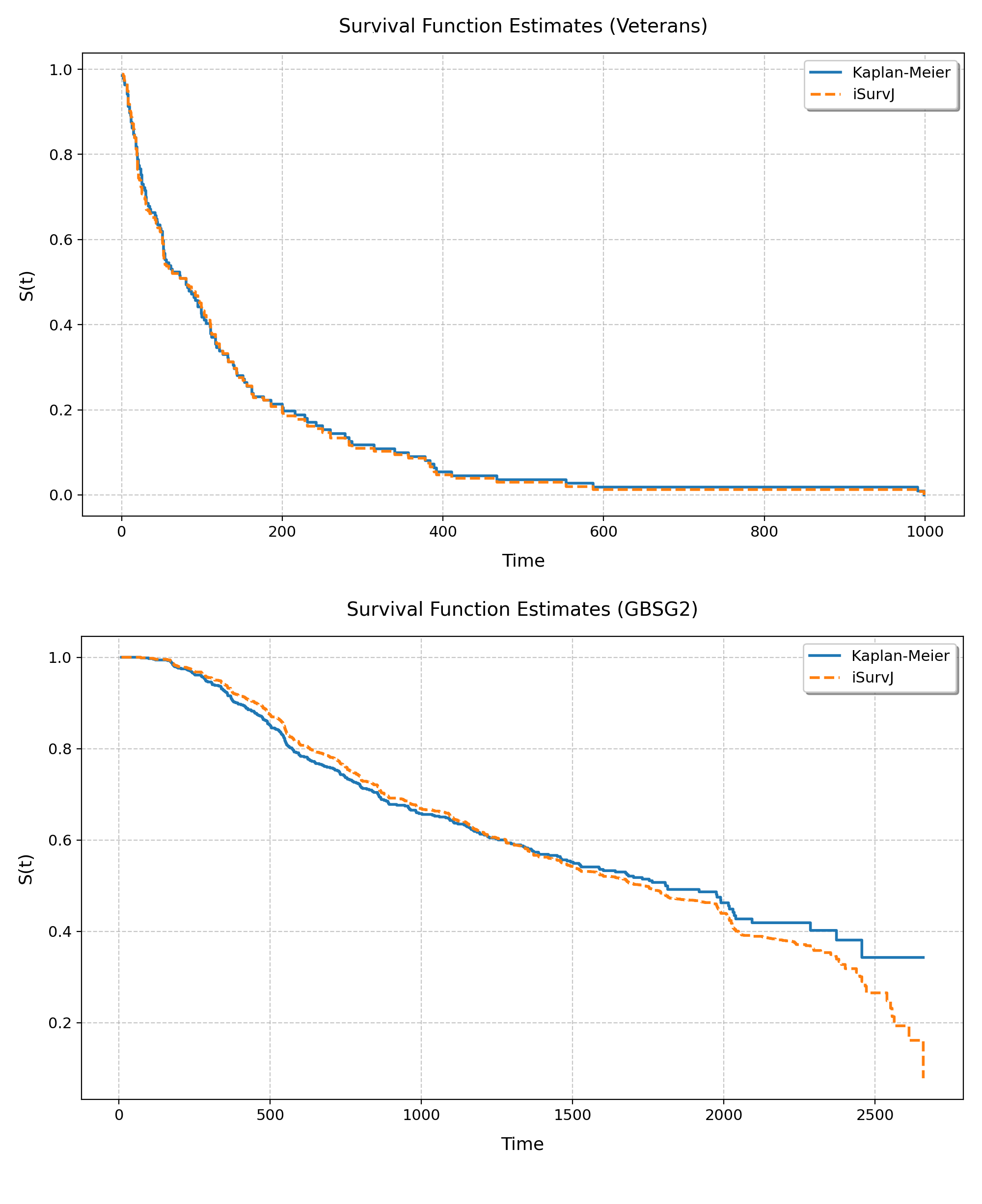}%
\caption{Unconditional SFs for the Veterans and GBSG2 datasets}%
\label{f:uncond_veterans_gbsg2}%
\end{center}
\end{figure}

\begin{figure}
[ptb]
\begin{center}
\includegraphics[
height=5.0052in,
width=4.1717in
]%
{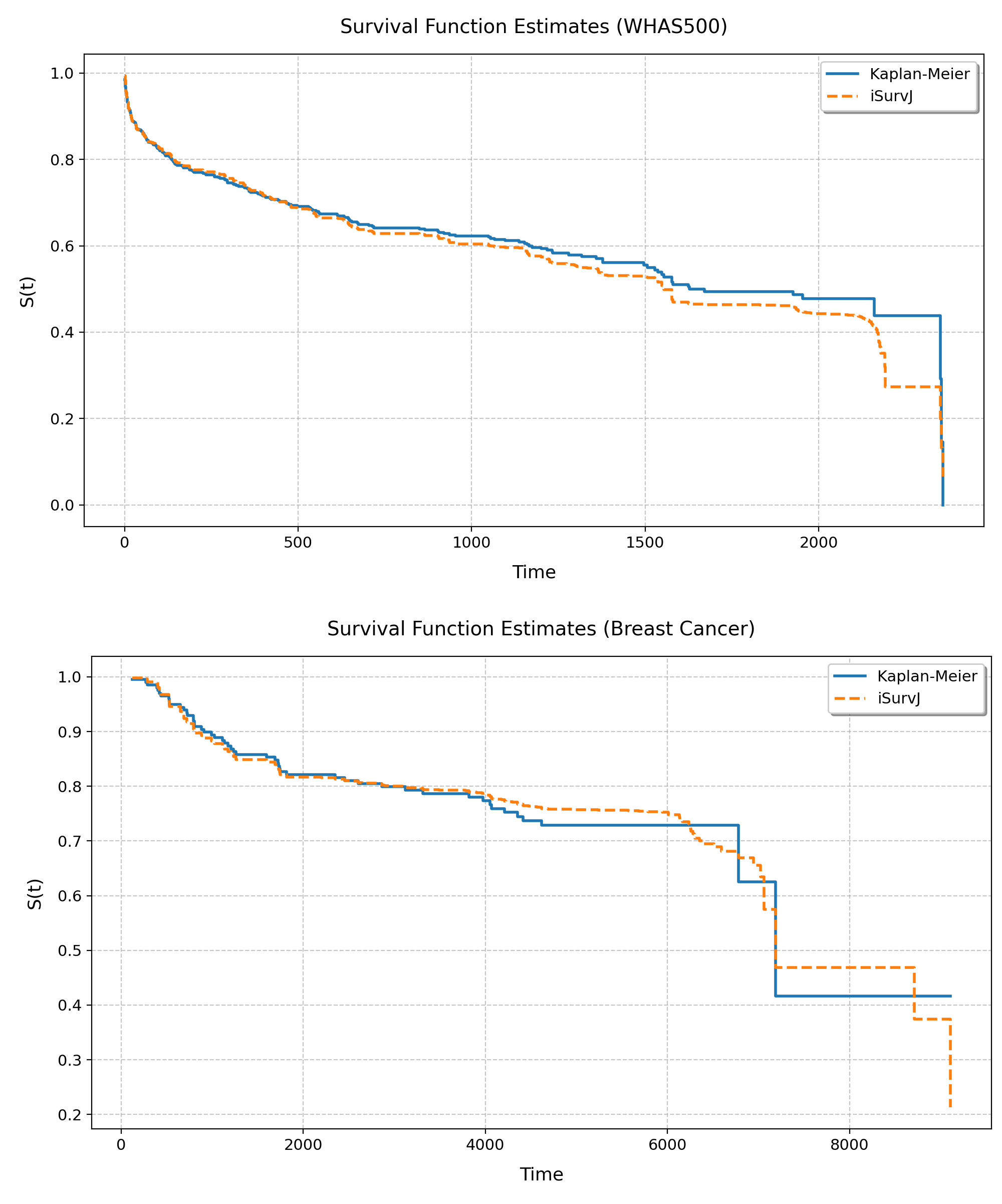}%
\caption{Unconditional SFs for the WHAS500 and Breast Cancer datasets}%
\label{f:uncond_whas500_bc}%
\end{center}
\end{figure}

Figs.~\ref{f:uncond_veterans_gbsg2} and~\ref{f:uncond_whas500_bc} illustrate
unconditional SFs obtained by iSurvJ and the Kaplan-Meier estimator. One can
see from the plots that SFs of two models are very close to each other.%

%

\subsubsection{An illustrative Comparison Example of iSurvJ(G) and Beran
estimator}

\hspace{2em}

To demonstrate the quality of the proposed approach, a synthetic dataset is
generated in which the true relationship between a feature and the time to
event is given by the following parabolic function:%

\[
T=(x-x_{0})^{2},
\]
where $x_{0}=0$ corresponds to the position of the minimum.

Features $x\in \mathbb{R}$ were generated to ensure a non-uniform observation
density in different intervals: $200$ points uniformly distributed in the
interval $[-5,-2]$, $10$ points located in the interval $[-2,2]$, $200$ points
uniformly distributed in the interval $[2,5]$. Thus, the total number of
observations is $N=410$, with the central region corresponding to the smallest
time-to-event values artificially thinned to reduce event density.

For the left and right parts of the sample, the censoring indicator $\delta
\in \{0,1\}^{n}$ is performed randomly according to the Bernoulli scheme with
probability $p_{\text{cens}}=0.1$. For the central part of the parabola
containing 10 observations, the censoring indicators are set manually,
alternately taking values $0$ and $1$, which ensured diversity of observations
in the region of the minimum. The iSurvJ(G) model is used in the experiment
with the number of epochs $50$, the learning rate $7$, $\gamma=0.2$, $k=5$,
the batch\_rate $1$. In the Beran estimator, the same kernel as in iSurvJ(G)
is used.

\begin{figure}
[ptb]
\begin{center}
\includegraphics[
height=2.5639in,
width=4.259in
]%
{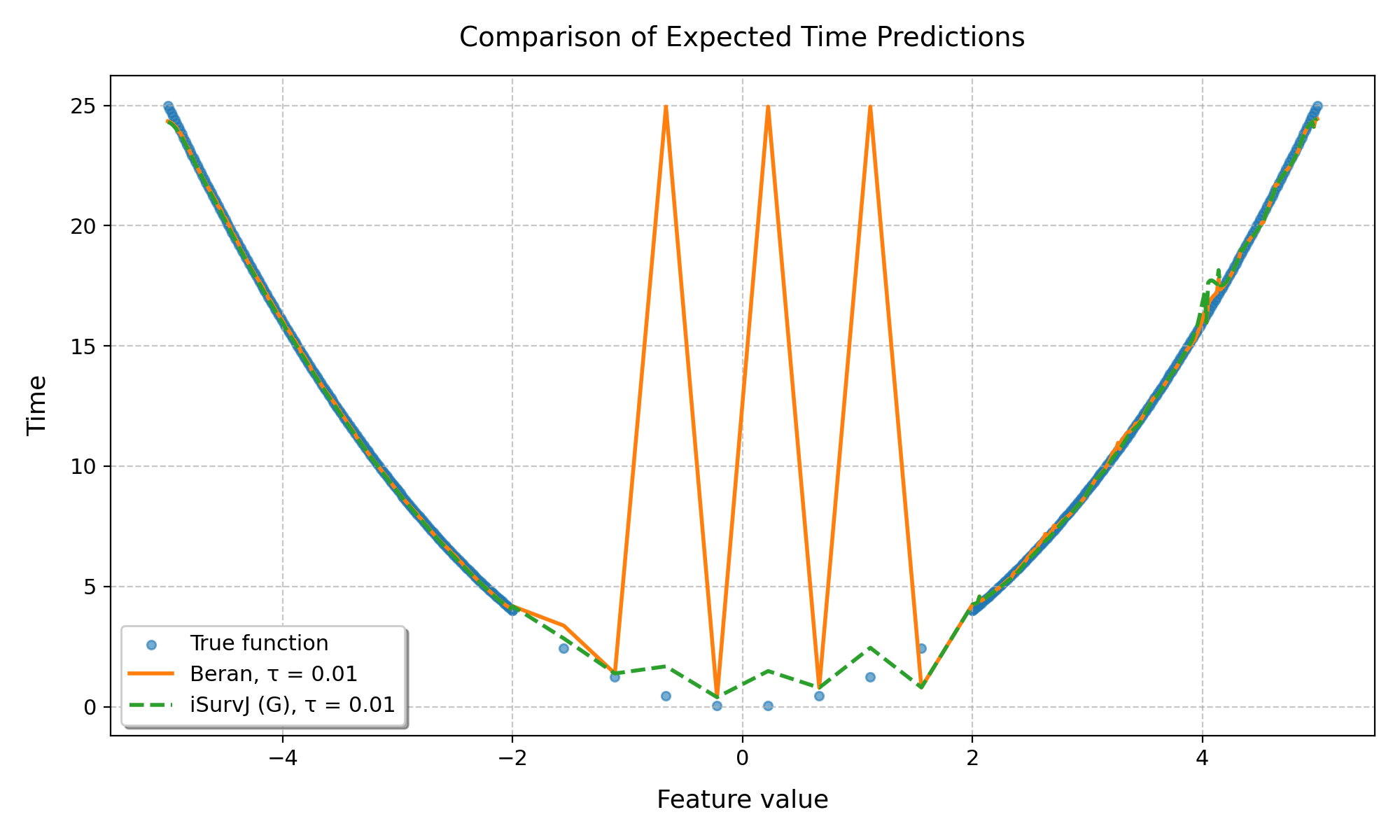}%
\caption{Comparison of predicted expected times obtained by the Beran
estimator and iSurvJ(G)}%
\label{f:parabola}%
\end{center}
\end{figure}

Results of the experiment are shown in Fig. \ref{f:parabola}, which clearly
demonstrates that the Beran estimator exhibits unstable behavior and fails to
learn effectively from the given data, while the proposed model iSurvJ(G) with
the same kernel successfully handles the complex data structure.%

\end{document}